\documentclass[sigconf]{acmart}

\usepackage{tabularray}
\usepackage{enumitem}
\usepackage{subcaption}
\usepackage{xspace}
\usepackage[frozencache]{minted}
\usepackage{soul}
\usepackage{wrapfig}
\usepackage{ifthen}
\DeclareEmphSequence{\bfseries}
\graphicspath{{./figures}}
\usepackage{enumitem}
\UseTblrLibrary{booktabs}
\setlist{nosep}
\usepackage{afterpage}
\usepackage{tikz}
\let\oldquote\quote
\let\oldendquote\endquote

\renewenvironment{quote}[1][]{
\def\speakername{#1}%
\oldquote
\itshape
``}{''
\normalshape
\hfill\ifthenelse{\equal{\speakername}{}}{}{(\speakername)}
\oldendquote
}
\newcommand{\inline}[2][]{
\textit{``#2''}\ifthenelse{\equal{#1}{}}{}{(#1)}
}

\DeclareRobustCommand{\projectname}{\textsc{UXAgent}\xspace}

\newmintinline[code]{text}{}

\setcopyright{acmlicensed}
\copyrightyear{2018}
\acmYear{2018}
\acmDOI{XXXXXXX.XXXXXXX}

\acmConference[Conference acronym 'XX]{Make sure to enter the correct
  conference title from your rights confirmation emai}{June 03--05,
  2018}{Woodstock, NY}
\acmISBN{978-1-4503-XXXX-X/18/06}

\begin{document}

\title{UXAgent: A System for Simulating Usability Testing of Web Design with LLM Agents}
\author{Yuxuan Lu}
\authornote{This work was done when Yuxuan was an intern, and Dakuo was a visiting scholar at Amazon. Contact email: \{\href{mailto:lu.yuxuan@northeastern.edu}{lu.yuxuan}, \href{mailto:d.wang@northeastern.edu}{d.wang}\}@northeastern.edu
}

\affiliation{%
  \institution{Northeastern University}
  \country{USA}
}

\author{Bingsheng Yao}
\affiliation{%
  \institution{Northeastern University}
  \country{USA}
}

\author{Hansu Gu}
\affiliation{%
  \institution{Amazon}
  \country{USA}
}

\author{Jing Huang}
\affiliation{%
  \institution{Amazon}
  \country{USA}
}

\author{Zheshen (Jessie) Wang}
\affiliation{%
  \institution{Amazon}
  \country{USA}
}

\author{Yang Li}
\affiliation{%
  \institution{Amazon}
  \country{USA}
}

\author{Jiri Gesi}
\affiliation{%
  \institution{Amazon}
  \country{USA}
}

\author{Qi He}
\affiliation{%
  \institution{Amazon}
  \country{USA}
}

\author{Toby Jia-Jun Li}
\affiliation{%
  \institution{University of Notre Dame}
  \country{USA}
}

\author{Dakuo Wang}
\affiliation{%
  \institution{Northeastern University}
  \country{USA}
}

\renewcommand{\shortauthors}{Lu et al.}
\setcopyright{none}
\begin{abstract}
Usability testing is a fundamental research method that user experience (UX) researchers use to evaluate and iterate their new designs. But what about \textbf{evaluating and iterating the usability testing study design itself}?
Recent advances in Large Language Model-simulated Agent (\textbf{LLM Agent}) research inspired us to design \textbf{\projectname} to support UX researchers in evaluating and iterating their study design before they conduct the real human-subject study.
Our system features a Persona Generator module, an LLM Agent module, and a Universal Browser Connector module to automatically generate thousands of simulated users and to interactively test the target website.
The system also provides a Result Viewer Interface so that the UX researchers can easily review and analyze the generated qualitative (e.g., agents' post-study surveys) and quantitative data (e.g., agents' interaction logs), or even interview agents directly. 
Through a heuristic evaluation with 16 UX researchers, participants praised the innovation of our system but also expressed concerns about the future of LLM Agent usage in UX studies.
\end{abstract}

\ccsdesc[500]{Human-centered computing~Empirical studies in HCI}
\ccsdesc[300]{Human-centered computing~Interactive systems and tools}

\keywords{Usability Testing, User Simulation, LLM Agent, Web Testing, UX}

\begin{teaserfigure}
    \centering
    \includegraphics[width=\linewidth]{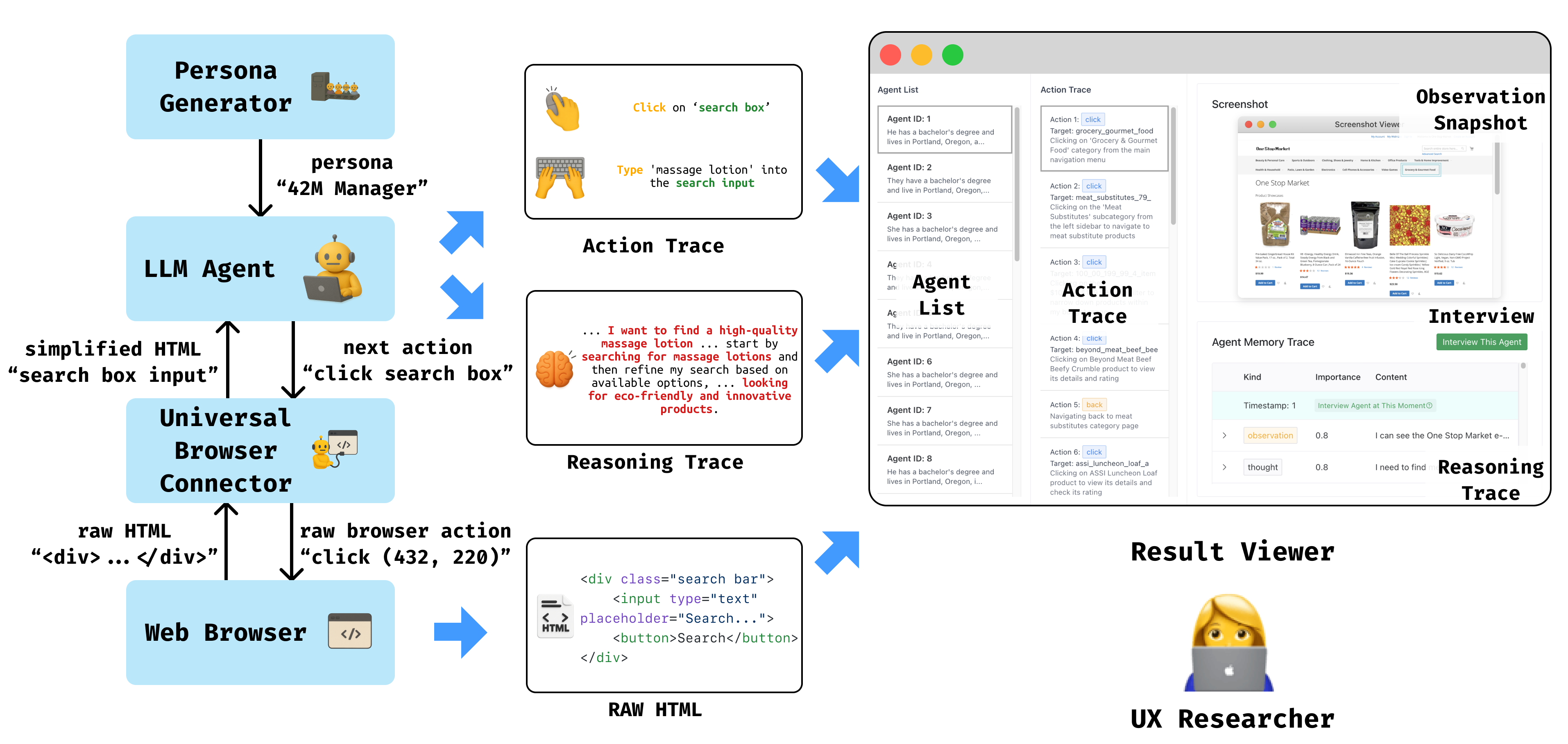}
    \caption{UXAgent is an open-sourced system. A UX researcher can use the Persona Generator to generate thousands of random personas and create LLM Agents with these personas. The LLM Agent then interacts with the target webpage design via a multi-turn interaction as if it were a user. Generated multimodal log data (e.g., Action Trace, Reasoning Trace, or Post-Study Survey) is presented to the UX researcher through a Result Viewer interface and can be exported as an Excel spreadsheet. UX researchers can also interactively interview these agents at any time point of a simulated UX session to mimic contextual inquiry or post-study interview UX methods.}
    \Description{A system diagram illustrating how a simulated UX study is conducted using the \projectname system. The process begins with the Persona Generator, which creates a user persona (e.g., “42M Manager”). This persona is passed to the LLM Agent, which interacts with a real Web Browser through a Universal Browser Connector. The connector converts raw HTML from the web page into simplified HTML for the agent and translates the agent’s actions into executable browser commands. The agent produces both an action trace (e.g., “Click on search box”) and a reasoning trace (e.g., thoughts on searching for massage lotion). These outputs are visualized in Result Viewer which enables UX researchers to review the action trace and chat with the agent and collect qualitative feedback. The UX researcher is shown in the center analyzing both the simulation and the agent's responses.}
    \label{fig:system-arch}
\end{teaserfigure}

\maketitle

\section{Introduction}

Usability testing is a fundamental research method for evaluating the user experience (UX) of a new feature or a new webpage design~\cite{shawHandbookUsabilityTesting1996a}.
Typically, a UX researcher (or a UI/UX designer, product manager, etc.) leads the design and execution of a usability testing study: they start with a new feature or web design, and then they need to come up with a UX study design (e.g., A/B testing and the user tasks), recruit participants, execute the study, monitor, and analyze the data.
However, the current process of conducting usability testing often faces challenges, especially in the study design stage and the study execution stage~\cite{folstadAnalysisPracticalUsability2012, hertzumEvaluatorEffectChilling2003, kuangMergingResultsNo2022, norgaardWhatUsabilityEvaluators2006}.
In the study design stage, researchers often focus on collecting early feedback on new feature design to iterate the feature, but the study design itself is left unevaluated~\cite{shawHandbookUsabilityTesting1996a}.
Study design flaws that surface only during execution can lead to costly consequences. This is especially critical when working with underrepresented groups (e.g., non-binary shoppers), where participants are a limited and valuable resource, thus the findings may not be significant or insightful.
In such cases, a UX practitioner in a company may not even have a second chance to iterate the study design.

To alleviate this issue, researchers have practiced using \textbf{pilot study} sessions~\cite{decker-maurerMethodMadnessUsability2012} or \textbf{empathy-based research} methods~\cite{sarkelaMethodEmpathyBasedStories2020}:
Researchers often conduct a limited number of pilot study sessions before the actual study execution to test validity and uncover design flaws. However, these sessions are time-consuming and financially costly, since their data are not considered valid test data and require additional participant recruitment~\cite{decker-maurerMethodMadnessUsability2012}.
Researchers sometimes also apply an empathy-based research method, during which they attempt to put themselves in the shoes of their target users, especially when designing and studying for an underrepresented user population~\cite{sarkelaMethodEmpathyBasedStories2020}.
However, this method is inherently limited, as researchers may fail to accurately represent the target population and may unintentionally introduce bias \cite{whartonApplyingCognitiveWalkthroughs1992}. 
These limitations highlight the need for novel methods and systems that can offer early feedback on the study design as well as on the feature designs to support UX researchers.

With the advancement of Large-Language Models (LLMs), some researchers have prompted LLMs to act as autonomous agents (i.e., \textbf{LLM Agent}) that can perceive the environment and generate human-like actions to interact with the environment \cite{wangSurveyLargeLanguage2024}. 
For example, researchers have successfully empowered LLM Agents to simulate a group of 25 human agents living in a village \cite{parkGenerativeAgentsInteractive2023},
replicate participants in a social science study \cite{parkGenerativeAgentSimulations2024,schmidgallAgentClinicMultimodalAgent2024,leeApplicationsGPTPolitical,gurcanLLMAugmentedAgentBasedModelling2024}, act as patients and clinicians in a hospital \cite{liAgentHospitalSimulacrum2024}, and behave as programmers for a software development company \cite{qianChatDevCommunicativeAgents2024}.
The advances in LLM agent research inspire us that: if we can simulate diverse user behaviors and perspectives with an LLM agent as surrogates for real human-subject participants, then these surrogates can interact with real-world web environments to produce rich usability testing data.
Such data can support UX researchers to analyze and identify potential flaws in their study design before conducting real user studies, as well as offer UX researchers early feedback for feature designs.

Recently, a few LLM Agent-based systems (e.g., WebAgent~\cite{gurRealWorldWebAgentPlanning2023}, WebVoyager~\cite{heWebVoyagerBuildingEndtoEnd2024}, etc.) have explored simulating human interaction behaviors with web-based systems.
For example, ReAct~\cite{yaoReActSynergizingReasoning2023} can generate both reasoning and action trace and simulate human customers on WebShop \cite{yaoWebShopScalableRealWorld2022}, a commonly used sandbox environment for e-commerce platforms.
Another example is Computer-Using Agent systems such as OpenAI Operator\footnote{\url{https://openai.com/index/introducing-operator/}} and Claude’s Computer-use API\footnote{\url{https://www.anthropic.com/news/3-5-models-and-computer-use}} that can complete tasks in computer environments.
However, these systems fall short of our use case: we need an LLM Agent to be able to operate with the dynamically changing web pages, to think (generate reasonings) as human users do, and to interact with the web page within a human-like responsive time interval.
The aforementioned LLM Agent systems (1) are restricted to a sandbox simulation instead of a real-world complex web system (e.g., ReAct~\cite{yaoReActSynergizingReasoning2023}), (2) rely only on complex and heavy reasoning architectures that hinder real-time responsiveness with the web (e.g., ~\cite{maLASERLLMAgent2024,heWebVoyagerBuildingEndtoEnd2024,gurRealWorldWebAgentPlanning2023}), (3) do not have the reasoning depth so the simulated behavior trace output is too simple in comparison to real-human's actions~\cite{sodhiStePStackedLLM2024}.
These challenges call for a new LLM Agent architecture design (Fig.~\ref{fig:structure}) that can balance the reasoning capabilities with interaction speed when meeting complex and diverse real-world web systems.

In this work, we propose \textbf{\projectname} (Fig.~\ref{fig:system-arch}), a system that can generate LLM Agents as usability testing participants at scale and run simulated interactions with a given real-world web environment to collect simulated user behavioral data.
The UX researcher can define a demographic distribution and use the Persona Generator to generate thousands of personas for the LLM Agents, and then use the Universal Browser Connector module to interact with webpages via Chrome. 
Inspired by \citet{parkGenerativeAgentsInteractive2023} and the Dual Process Theory \cite{kahneman2011thinking,evansDualProcessTheoriesHigher2013,zhangLeveragingDualProcess2025,christakopoulouAgentsThinkingFast2024}, \projectname includes \textbf{an LLM Agent module }with a new LLM Agent architecture to have a two-loop structure to support both in-depth reasoning and real-time response.
Our \textbf{Universal Browser Connector modul}e allows LLM Agents to seamlessly parse and interpret real-world webpages while automatically executing real actions on the webpage without the need for manually pre-defined action spaces.
\projectname also provides a \textbf{Result Viewer Interface} that enables UX researchers to seamlessly review LLM agents’ interactions with web pages, examine post-study survey results, and conduct qualitative interviews with the agents.
Our position is that: \textbf{LLM agents are not meant to replace human participants in UX studies, but rather to help UX researchers to iteratively revise the study design, thus to be more responsible to human participants}.
LLM Agents can work together with UX researchers (human-AI collaboration~\cite{wangHumanAICollaborationData2019}) in a simulated pilot study manner to provide the desired early and immediate feedback for UX researchers to further iterate their design before testing with real human participants.

To evaluate our system, we conducted a user study with 16 UX researchers and asked them to analyze usability testing data (action trace, reasoning trace, and qualitative interview) of 20 LLM Agents interacting with a shopping website, as if they were the study owners.
Our study scenario is that: a UX researcher is tasked to design a usability testing study to gather feedback for a newly designed shopping website feature (product filter menu) and propose new designs for this website feature.
Our results suggest that the UX researcher participants were able to review and analyze the LLM-agent participants' data generated by \projectname, identify potential study design flaws, and come up with system feature improvement ideas. In comparison to the initial UX study design, they were satisfied with their revised study designs after analyzing the data, despite some criticizing that the LLM agent's generated actions were not what real users would do.
We conclude our paper by discussing how LLM agents may play an increasingly important role in the future of UX research.

\section{Related Work}
\subsection{Challenges in Usability Testing}
Usability testing is a core component of UX research, used to evaluate how easily users can interact with a product to achieve their goals \cite{shawHandbookUsabilityTesting1996a,barnum2020usability, bastienUsabilityTestingReview2010, lewis2012usability}.
It involves observing real users as they navigate through tasks and providing valuable feedback on design effectiveness \cite{barnum2020usability}. 
This method helps refine products by identifying usability issues, measuring user satisfaction, and improving overall user experience. The key benefits of usability testing include validating design choices, avoiding internal biases, and ensuring the product meets user expectations.

However, UX researchers conducting usability testing on web designs have been facing multiple challenges in the \textbf{study design stage} and the \textbf{study execution stage}~\cite{folstadAnalysisPracticalUsability2012, hertzumEvaluatorEffectChilling2003, kuangMergingResultsNo2022, norgaardWhatUsabilityEvaluators2006}.  
During the study design stage, it is often difficult for researchers to craft appropriate study protocols and gather early feedback to iteratively improve the study design.  
To address this challenge, researchers have proposed guidelines and frameworks to support the design of usability testing, including methods for refining interview protocols~\cite{castillo2016preparing, shoozan2024application, roberts2020qualitative} and developing survey questionnaire items~\cite{rasinski2008designing, muller2015designing, muller2014survey, koelle2020social}.
To further streamline this process, recent work has begun to explore automated approaches for refining study protocols. For example, \citet{calvanoLeveragingLargeLanguage2025} investigated the use of Large Language Models (LLMs) to assist in generating components of usability studies, such as user tasks and evaluation criteria.  
To obtain early feedback on their designs, researchers typically rely on pilot studies, which, while valuable~\cite{hertzog2008considerations, truong2017pilot, van2001importance, oppenlaender2024state}, are time-consuming, costly, and limited in scale~\cite{Janaka2024, oppenlaender2021hardhats}.  
Despite these efforts, there remains a lack of efficient and scalable methods for UX researchers to evaluate and refine study designs before conducting full-scale studies with human participants.

For the study execution stages, a key challenge lies in the high cost and effort required to recruit enough qualified participants~\cite{Nebeling2013, Jane2024}. To mitigate these challenges, researchers have begun exploring the use of agents to simulate real human participants in usability testing~\cite{ren2014agent}. 
Prior work has introduced agents in various domains, including GUI testing~\cite{eskonenAutomatingGUITesting2020}, game environments~\cite{stahlkeArtificialPlayfulnessTool2019,fernandesAgentsAutomatedUser2021}, and accessibility testing~\cite{taebAXNavReplayingAccessibility2024, zhong2025screenaudit}. 
A particularly notable effort is SimUser~\cite{xiangSimUserGeneratingUsability2024}, which leveraged LLM Agents to simulate usability testing for mobile applications. In their approach, a user agent simulated human interaction, while an application agent represented the mobile app. These two agents interacted during the simulation to generate behavioral data. However, because the application under test was also simulated, their approach was limited in its ability to evaluate real-world interfaces and behaviors.
Despite growing interest in agent-based usability testing, the application of agents to real-world web design usability testing remains underexplored—largely due to the complexity of real web environments and the limitations of traditional web automation technologies.

\subsection{Simulation of Human Behavior with LLM and LLM Web Agents}

In the field of HCI, recent studies have demonstrated that LLMs possess the capability to simulate human-like behaviors, enabling research on personalized agent behavior. Unlike task-oriented agents, which are typically focused on achieving narrowly defined goals, \textbf{human agents} or \textbf{role-play agents}~\cite{chen2025towards} are designed with rich personas that encompass not only their roles and expertise but also their preferences, habits, and subjective viewpoints. 
For example, \citet{parkGenerativeAgentsInteractive2023} created a simulated town inhabited by 25 LLM Agents, each with a unique persona, and showed that these agents could exhibit believable and autonomous human-like interactions. \citet{parkGenerativeAgentSimulations2024} extended this line of work by simulating responses from 1,052 individuals in a social science survey, achieving an accuracy rate of 85\%.
Similarly, \citet{luPromptingNotAll2025} explored the use of LLM Agents to simulate virtual customers’ shopping behaviors on e-commerce platforms. 

LLM Agents have also been applied to study user attitudes toward privacy. For instance, \citet{zhangPrivacyLeakageOvershadowed2025,chenEmpathyBasedSandboxApproach2024} examined how agents with customizable personas can generate realistic search histories and profile behaviors to explore privacy concerns. In accessibility research, \citet{taebAXNavReplayingAccessibility2024} proposed AXNav, which uses LLMs to convert manual test instructions into replayable navigation videos for evaluating interface accessibility.
These emerging applications inform our approach: we aim to simulate realistic user behavior in web environments for usability testing, leveraging LLM Agents with diverse, persona-driven characteristics.

Recent advances in LLM-based Web Agents have significantly enhanced the ability of LLMs to assist with web-based tasks.
Early work like WebGPT~\cite{nakanoWebGPTBrowserassistedQuestionanswering2022} enabled GPT models to interact with search engines, improving question-answering performance. Subsequent systems, including WebVoyager~\cite{heWebVoyagerBuildingEndtoEnd2024}, LASER~\cite{maLASERLLMAgent2024}, and WebAgent~\cite{gurRealWorldWebAgentPlanning2023}, extended LLM capabilities to multimodal interaction, complex state space navigation, and long-horizon planning. Claude’s Computer-use API\footnote{\url{https://www.anthropic.com/news/3-5-models-and-computer-use}} further demonstrated precise, general-purpose control over user interfaces beyond web browsers.

Building on top of these advances, researchers have begun to explore the simulation of \textbf{human online behavior}. \citet{luPromptingNotAll2025} proposed a fine-tuning method to enhance model accuracy in predicting the next human action. OPERA~\cite{wangOPeRADatasetObservation2025} introduced a dataset of user online shopping behavior and benchmarked existing LLMs. Inspired by recent advancements in reasoning-enabled LLMs, Shop-R1~\cite{zhangShopR1RewardingLLMs2025} explored enhancing performance through RL. Still, existing work in the NLP field largely focuses on accuracy rather than mimicking human thinking processes. Since these approaches cannot capture qualitative aspects of user decision-making, key challenges in usability testing remain unresolved.

\subsection{Reasoning Architecture of LLM Agents}

Human cognition is often described as operating through two distinct systems: System 1 and System 2~\cite{kahneman2011thinking}.
System 1 governs fast, intuitive, emotional, and automatic responses, allowing humans to react quickly with minimal effort.
In contrast, System 2 is responsible for slower, more deliberate processes: it involves conscious reasoning, logical analysis, and problem-solving~\cite{kahneman2011thinking,evansDualProcessTheoriesHigher2013}.
Inspired by this dual-process theory, researchers have explored a variety of reasoning architectures in LLM Agent systems to balance speed and depth of thought.

For real-time responsiveness, many systems use System 1-style reasoning with single-prompt agents like OpenAI Operator\footnote{\url{https://openai.com/index/introducing-operator/}} and Claude’s Computer-use API\footnote{\url{https://www.anthropic.com/news/3-5-models-and-computer-use}}, which act on immediate inputs without deeper reflection. However, such agents often fail on complex tasks; for example, \citet{luPromptingNotAll2025} found System 1 alone insufficient for simulating realistic user behavior. To improve reasoning, researchers have introduced System 2-style approaches, such as chain-of-thought~\cite{weiChainofThoughtPromptingElicits2023}, ReAct~\cite{yaoReActSynergizingReasoning2023}, and reinforcement learning–based fine-tuning~\cite{deepseek-aiDeepSeekR1IncentivizingReasoning2025}.
These models excel in complex domains like programming and solving olympiad-level math problems~\cite{deepseek-aiDeepSeekR1IncentivizingReasoning2025,sunChallengingBoundariesReasoning2025}, but their high computational cost and latency~\cite{zhouSurveyEfficientInference2024,huaInteractiveSpeculativePlanning2024} limit their usability in real-time applications like interactive usability testing.

More recently, researchers have sought to integrate the strengths of both systems into unified reasoning architectures that can support real-time applications while preserving in-depth reasoning.
For instance, \citet{christakopoulouAgentsThinkingFast2024} introduced a talker-reasoner framework for a sleep coaching agent, simulating the interplay between fast reactions and deeper reflections.
Similarly, \citet{zhangLeveragingDualProcess2025} proposed DPT-Agent, which combines System 1 and System 2 thinking for efficient, simultaneous human-AI collaboration in game settings.
In particular, DPT-Agent uses System 2 to generate high-level instructions, while System 1 is used to translate those instructions into executable policy code for a finite-state machine.
However, extending this approach to the usability study scenario with web environments poses new challenges. 
Unlike structured game environments, web environments are dynamic and unpredictable -- each click can drastically change the context.
Agents need to continuously adapt their plans in response to newly observed content.
Building on these insights, we propose a two-loop architecture that combines fast, low-latency responses with reflective, high-level reasoning to simulate realistic, goal-directed user behavior in real-world web environments.

\section{\projectname: LLM-Agent-Based System for Usability Testing}

In this project, we propose \projectname, a system that can generate LLM Agents as usability testing participants at scale and run simulated interactions with a given web environment to collect simulated user behavioral data.
The architecture of the \projectname{} is illustrated in Figure~\ref{fig:system-arch}. Our system comprises four key components.
The \textbf{Persona Generator Module} generates a large-scale set of diverse personas, which are fed into the \textbf{LLM Agent}. 
The \textbf{LLM Agent Module} interacts with a web browser through the \textbf{Universal Browser Connector Module}.
The \textbf{Universal Browser Connector Module} parses the raw HTML as observation and robustly executes agent actions on the web page.
The \textbf{Result Viewer} allows the UX researcher users to easily review and analyze the generated qualitative (e.g., agents' post-study surveys) and quantitative data (e.g., agents' interaction logs), or even interview agents directly. 

\subsection{Persona Generator}

To support large-scale simulations with users from diverse backgrounds, our system features a Persona Generator module that automatically creates a wide range of agent personas.  As an example, generating 1,000 personas takes about two mins.
Users begin by specifying an example persona (e.g., mid-career professional marketing manager; a full example can be found in Appendix \ref{sec:example-persona}) and a target demographic distribution (e.g. 10\% low-income, etc.).

The system uses the input example persona as a seed to generate the desired number of agent personas. It first samples demographic attributes based on the specified distribution and then builds full personas using those attributes. To enhance diversity, the system randomly selects one of the previously generated personas to serve as the example in the LLM prompt for generating each new persona.
This prevents repetitive outputs, encourages variation in traits and backgrounds across the generated personas, and allows the system to efficiently generate a rich, demographically balanced set of personas.
The prompts used for persona generation are detailed in Appendix \ref{sec:prompt-persona}.

\subsection{LLM Agent Architecture for Usability Testing}

\subsubsection{Reasoning and Planning Architecture}
Inspired by the Dual Process Theory~\cite{kahneman2011thinking,evansDualProcessTheoriesHigher2013}, we designed the LLM Agent with a two-loop structure comprising a \textbf{Fast Loop} and a \textbf{Slow Loop} (Figure~\ref{fig:structure}). 
The Fast Loop enables \textbf{rapid interactions with minimal latency}, allowing the agent to promptly respond to changes in the web environment and support real-time simulation.
In contrast, the Slow Loop performs \textbf{in-depth reasoning}, generating high-level insights and strategic guidance to ensure that the agent's behavior aligns with its intent and persona.
Each loop consists of modules that retrieve memory pieces from the \textbf{Memory Stream} (Section~\ref{sec:memory-stream}) and generate memory pieces to be stored in the Memory Stream.
The Perception Module produces \textit{observations}, the Planning Module generates \textit{plans}, the Action Module records \textit{actions}, and the Wondering and Reflection Modules create \textit{thoughts}.

The Fast Loop and Slow Loop operate \textbf{in parallel}, and share information via the \textbf{Memory Stream}.
For example, high-level reflections generated by the Slow Loop can influence subsequent planning and actions in the Fast Loop. 
This non-blocking, asynchronous architecture allows the agent to reason deeply while still maintaining responsiveness, making it well-suited for real-time simulations in complex web environments.

\definecolor{fastloop}{HTML}{005EA6}
\definecolor{slowloop}{HTML}{F37E24}
\definecolor{module}{HTML}{E1708F}
\begin{figure}[t]
    \centering
    \includegraphics[width=.8\linewidth]{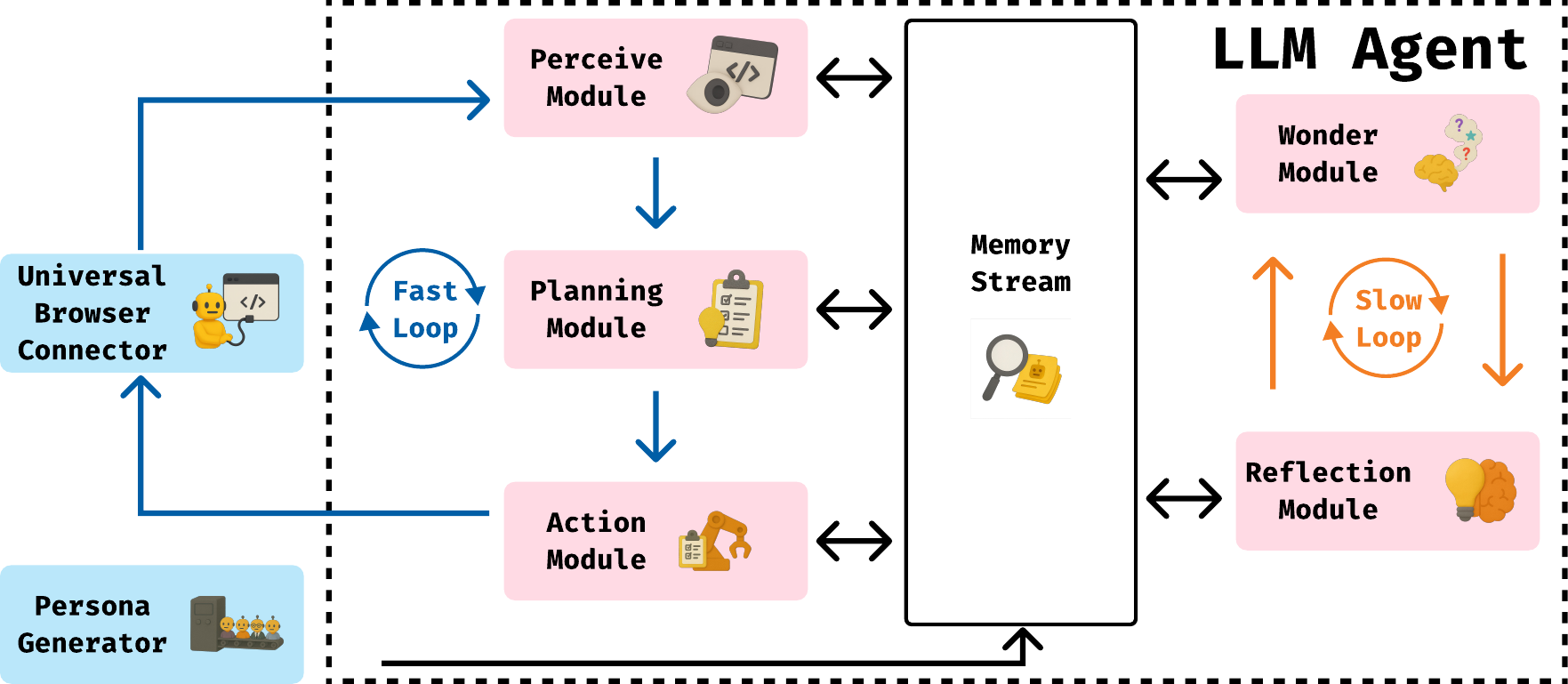}
    \caption{LLM Agent Architecture design.
    The architecture consists of two loops, the \textcolor{fastloop}{Fast Loop} and the \textcolor{slowloop}{Slow Loop}, responsible for fast interaction with the web and for in-depth thinking, respectively. In the fast loop, the perception module translates the raw observation from the browser connector to natural language observation memories that are stored in the memory stream. The planning module makes short-term plan that is executed by the action module.
    In the slow loop, the wonder module mimics humans' ``mind drift'' and the reflection module makes in-depth reasoning and reflection about the current situation.
    Memories stored in the memory stream are retrieved by different modules, and each module will produce some memories that are written back to the memory stream.
    }
    \Description{A system diagram of the reasoning architecture for UXAgent, illustrating its two-loop design. On the left, inputs flow from the Persona Generator and the Universal Browser Connector into the LLM Agent. Inside the agent, there are two reasoning loops: the Fast Loop and the Slow Loop. The Fast Loop includes the Perceive Module (which processes the web environment), the Planning Module (which generates a plan based on observations), and the Action Module (which executes actions). These modules interact with a central Memory Stream that stores all memory pieces. The Slow Loop consists of the Reflection Module (which produces strategic insights) and the Wonder Module (which simulates spontaneous thoughts). Both loops access and contribute to the Memory Stream, enabling the agent to reason and act in a human-like manner in real time. Arrows indicate data flow and loop cycles, emphasizing the modular, asynchronous architecture of UXAgent.}
    \label{fig:structure}
\end{figure}

\begin{figure}[t]
    \includegraphics[width=\linewidth]{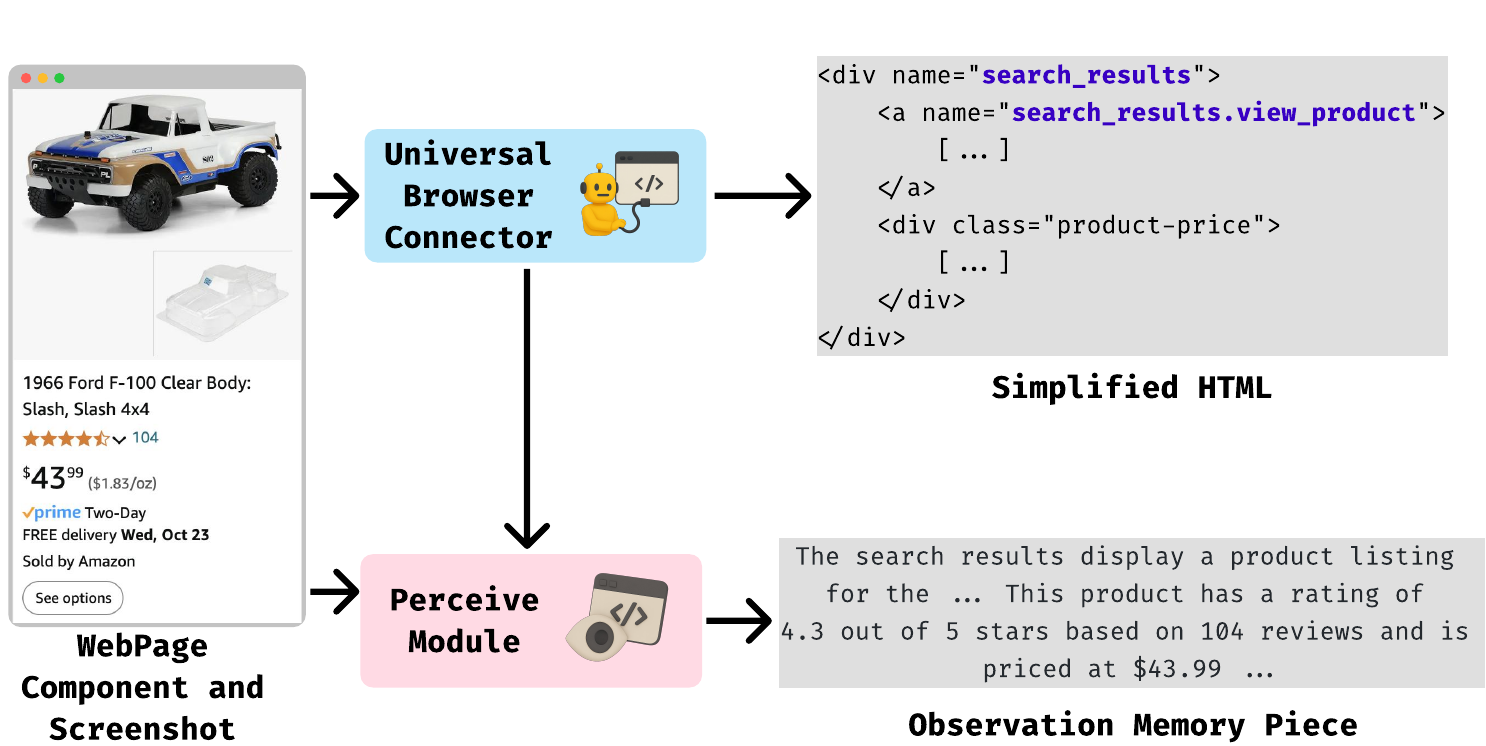}
    \caption{Example of a web page component processed by the Universal Web Connector and the Perception Module.
    The Universal Web Connector parses the raw web page (left) into a simplified HTML structure (upper right) and adds a semantically meaningful ID attribute to every interactable element. The Perception Module generates a descriptive observation memory (lower right) based on the simplified HTML and the screenshot.
    }
    \Description{A flow diagram illustrating how UXAgent converts a web page component into an observation memory piece. On the left, a product listing from an e-commerce website (showing a Ford truck body, price, rating, and delivery info) is processed by the Universal Browser Connector. The connector outputs simplified HTML code, shown in the top right, which contains structured and labeled tags like <div name="search_results">. Simultaneously, the same input is processed by the Perceive Module (bottom center), which transforms the product information into a natural-language summary called an "Observation Memory Piece." The text describes the product’s name, rating, number of reviews, and price, demonstrating how UXAgent interprets visual web content into structured, human-readable memories for reasoning.}
    \label{fig:perception-demo}
\end{figure}

\paragraph{Perception Module}
\label{sec:perception-module}
To allow the agent to observe and comprehend the outside environment, we designed the Perception Module.
The Perception Module observes the web environment and generates a series of memories describing the LLM Agent's observations.
Unlike simulated environments, our system interacts with real web pages, and the perception process is designed to mimic how a real human user might visually comprehend a web page.
As shown in Figure~\ref{fig:perception-demo}, the module systematically reads the web page from top to bottom.
The web page is then transformed into a text-form observation memory piece and stored in the Memory Stream.
These observation memories in the Memory Stream are then retrieved and used by other modules to inform subsequent planning, reasoning, and action.

\paragraph{Planning Module} 
\label{sec:planning-module}
To enable real-time decision-making, we designed the Planning Module to mimic human System 1–style rapid cognition by leveraging memory pieces generated by other modules.
This module continuously creates and updates short-term plans aligned with the agent’s intent and persona, based on relevant memories (such as observations and reflections) retrieved from the Memory Stream.
By relying on precomputed insights rather than performing explicit multi-step reasoning during execution, the Planning Module allows the agent to remain responsive while still benefiting from deeper, strategic thinking in the background.

The module mirrors human planning processes by breaking down tasks into logical, interpretable steps. For example, if the agent's initial intent is to ``buy a sofa,'' its plan may involve steps such as searching for sofas, applying filters for delivery date and price, comparing product details, and proceeding to purchase.
These plans are continuously refined as new information becomes available through perception and memory updates.
The generated plan is then stored as a plan memory piece in the Memory Stream, to be retrieved and used by other modules.

Specifically, the module performs three core functions:
\begin{itemize}
    \item \textbf{Initial Plan Creation:} At initialization, the agent formulates an initial plan based on the persona's intent. For example, if the intent is ``buy a sofa,'' the initial plan may involve searching online stores, reviewing search results, and selecting potential products.
    \item \textbf{Plan Updates:} As new observations and memories emerge, the module updates the plan accordingly. For instance, if the agent is browsing search results and finds a sofa that matches the persona's preferences (e.g., color or size), the plan evolves to include reviewing detailed information and considering a purchase.
    \item \textbf{Rationale and Next Step:} To ensure transparency and interpretability, the module provides a clear rationale for its current decisions and explicitly identifies the next step required to achieve the agent's intent, ensuring logical progression toward goals.
\end{itemize}

\paragraph{Action Module}
\label{sec:action-module}

The Action Module is responsible for executing the agent's planned actions in the web environment. It is designed to mimic how humans translate intentions into interface-level interactions. The module takes as input the current environment state, the agent’s persona and intent, the plan generated by the Planning Module, and relevant memories retrieved from the Memory Stream.

From the current step in the plan (e.g., ``I'm going to search for massage lotion''), the Action Module generates a corresponding low-level action defined in the predefined action space (see Section~\ref{sec:action-space}) as well as it's textual description.
The generated action is executed using the Web Browser Connector and also stored in the Memory Stream as an action memory piece.
For instance, the above plan step may result in an action like \code[breaklines]{type(``massage lotion'', search_input)}.
This design enables the agent to perform realistic, human-like interactions with real web environments, such as typing, clicking, navigating, and submitting forms.

\paragraph{Reflection Module}
\label{sec:reflection-module}
The Reflection Module simulates human reflective thinking by synthesizing recent memories—including observations, actions, previous plans, and reflections. It generates high-level insights that capture key takeaways and strategies,  mimicking how a person would reason in similar situations.
These reflections are stored in the Memory Stream as thought memory pieces, informing and guiding future decisions by the agent.

\paragraph{Wonder Module}
\label{sec:wonder-module}
Humans frequently engage in spontaneous, unrelated thoughts, allowing their minds to drift beyond the current task. To emulate this behavior, the Wonder Module prompts the agent to generate unrelated, spontaneous thoughts or ``wonders'' based on the persona and recent memories. Although these thoughts may not directly impact immediate tasks, they enhance realism in simulated user behavior.
Generated wonders are stored in the Memory Stream as thought memory pieces, becoming accessible to other modules.

\subsubsection{Memory Stream}
\label{sec:memory-stream}

The \textbf{Memory Stream} stores the agent's memory pieces (including observations, actions, reflections, and spontaneous thoughts) as natural-language descriptions with associated timestamps.
It serves as a critical bridge between the Fast Loop and the Slow Loop and enables these two asynchronous loops to exchange information and remain contextually aligned.

As memories accumulate, the Memory Stream becomes increasingly lengthy, posing efficiency challenges for memory retrieval.
To address this, we implement a retrieval module that prioritizes and retrieves memories relevant to the agent's current task.
Following prior work \cite{parkGenerativeAgentsInteractive2023}, our retrieval mechanism scores memories based on three dimensions: \textbf{importance}, \textbf{relevance}, and \textbf{recency}. The final retrieval score for each memory is calculated as:
\[
score = (importance \times w_{imp} + relevance \times w_{rel} + recency \times w_{rec}) \times w_{type}
\]
where $w_{imp}$, $w_{rel}$, and $w_{rec}$ represent the weights for importance, relevance, and recency, respectively, and $w_{type}$ is a weight determined by the memory type.
Each module uses separate parameter sets to align with their distinct retrieval priorities: modules in the Fast Loop emphasize recency for rapid actions, while modules in the Slow Loop emphasize relevance for deeper reasoning, mirroring human cognitive patterns.

\paragraph{Importance}
\label{sec:importance}

The importance score measures how closely a memory piece aligns with the agent's original intent. In real-world web environments, users often ignore irrelevant elements like ads or unrelated content. To mimic this behavior and improve simulation efficiency, each memory piece is assigned an importance score by an LLM, helping the agent prioritize task-relevant information (e.g., product listings, prices) over peripheral details. To maintain responsiveness, importance scores are computed only in the Slow Loop; memory pieces without a score are left undefined, and their importance scores are ignored during retrieval.

\paragraph{Relevance}
The relevance score reflects the semantic similarity between a memory and the current context. Each memory's natural-language description is transformed into a vector embedding, and its cosine similarity is calculated against a query string generated by the retrieval module.

\paragraph{Recency}  
The recency score prioritizes recently created memory pieces, as they are more likely to be relevant to the agent’s immediate task. Since our agent focuses on short-term, in-session simulation (as opposed to prior work that models long-term behavior) recency plays a critical role in supporting rapid, context-aware decision-making. It is computed using an exponential decay function based on memory timestamps:
\[
    \text{recency} = \frac{1}{e^{k(t_0 - t)}}
\]
where $t$ is the timestamp of the memory, $t_0$ is the current time, and $k$ is a decay rate constant. A higher $k$ makes the agent focus more heavily on the most recent memories, while a lower $k$ allows for a broader temporal range. Through empirical trial runs, we set $k = 1$ in our implementation.

\subsubsection{LLM Agent Generated UX Study Data Traces}  
\label{sec:data-collected}

To enable UX researchers to analyze both what the agent did and why --- thus supporting comprehensive evaluations of web interfaces from multiple perspectives --- \projectname generates a rich set of behavioral and cognitive data from the LLM agent to interact with the web environments via \textbf{ a multi-turn }interaction simulation. 

This data provides both quantitative and qualitative insights into the agent’s interaction with the web environment (Figure~\ref{fig:system-arch}). Specifically, it includes:

\textbf{Action Trace.} The system records every action generated by the Action Module in sequence. This trace captures how the agent navigates and manipulates the web interface, including clicks, form submissions, and navigations. It reflects the observable behavior of the agent and can be used for replay, analysis, or visualization of its interaction flow.

\textbf{Reasoning Trace.} In addition to observable actions, the system captures the agent’s internal reasoning trace, including all memory entries generated by the modules within the Reasoning Architecture. This includes observations, plans, rationales, reflections, and wonders. These memories 
mimic human participants' ``think-aloud'' sessions, provide a window into the agent’s decision-making process, and can be used to gather qualitative feedback from the agent.

\textbf{Post-Study Survey.} After completing their interaction, agents are asked to answer a pre-defined post-study survey. Loaded with their reasoning trace, they respond as if they were human participants, offering feedback on the system and their experience.

\subsection{Universal Web Connector}

We design a \textbf{Universal Web Connector} that enables LLM agents to \textbf{interact with any real-world web environments}. It provides a structured observation space derived from the browser’s DOM through our \textbf{DOM parser}, and robustly executes actions on the page. The interfaces are human-aligned: the observation space reflects how users perceive a web page, and the action space mirrors how users interact with it. Because the connector requires no per-site configuration, UX researchers can easily test new web designs without additional setup.

\subsubsection{Observation Space}
\label{sec:html-parser}
The observation space of the Universal Web Connector is structured as a JSON object containing:

\begin{itemize}
    \item \code{html}: Simplified HTML representation of the webpage.
    \item \code{clickable_elements}: List of clickable elements (e.g., buttons, links).
    \item \code{input_elements}: List of input elements (e.g., text fields, dropdowns).
    \item \code{hoverable_elements}: List of hoverable elements (e.g., hover-triggered menus, tooltips)
    \item \code{select_elements}: List of drop-down single select forms
    \item \code{tabs}: List of currently available browser tabs
    \item \code{error}: Description of errors encountered, such as interacting with a non-existent element.
\end{itemize}

To mimic human perception of a web page, we design a \textbf{DOM (Document Object Model) parser} that reduces the page to elements visible and meaningful to users. Invisible content such as scripts, styles, and metadata is discarded, along with elements hidden by CSS, rendered at zero size, or pushed entirely off-screen. To reduce structural redundancy, the parser collapses trivial nesting of non-semantic containers, such as chains of \code{<div>} wrappers, and prunes empty elements. Exceptions are made for controls that may legitimately appear empty (e.g., \code{<input>}, \code{<img>}, etc.). 

The parser heuristically determines whether elements are interactive. Nodes are marked as clickable if they are native controls (e.g., \code{<button>}, \code{<input>}), anchors with \code{href}, elements with explicit \code{onclick} handlers, or elements with ARIA roles such as \code{button} or \code{link}. Humans can easily distinguish between selectable text (indicated by ``\raisebox{-0.2ex}{\includegraphics[height=1.4ex]{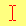}}'') and clickable elements (indicated by ``\raisebox{-0.2ex}{\includegraphics[height=1.4ex]{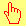}}''), whereas agents often attempt to click on plain text or other non-interactive elements. To reduce such errors, we expose the same cues available to humans: elements with a computed cursor style of \code{pointer} are considered clickable. To further identify hover-sensitive targets, we monkey-patch \code{addEventListener} so that whenever a node registers a hover event listener, it is annotated with \code{maybe-hoverable=true}.

To enable the model to reference elements during actions, each interactive element is assigned a stable, human-readable identifier derived from visible text, placeholders, or tag names. Identifiers are normalized to short slugs, scoped hierarchically, and made unique with numeric suffixes. Both the stripped node and original DOM share the same \code{semantic-id}, with clickable elements additionally marked \code{clickable=true}. Additionally, since JavaScript states such as input values are not directly available in the HTML, the parser extracts and records them, including values, selections, and focus, and assigns semantic identifiers where needed.

Optionally, when the user wants to evaluate layout-level features, such as comparing two different designs, the observation space can also include a full-page \textbf{screenshot of the webpage}. This provides access to visual cues such as spatial arrangement, alignment, and whitespace that are not recoverable from the stripped DOM alone. By default, the system uses text-only observations, which are more compact and faster to process, while the visual extension offers richer context at the cost of additional computation.

\begin{figure*}[t]
    \centering
    \includegraphics[width=1\linewidth]{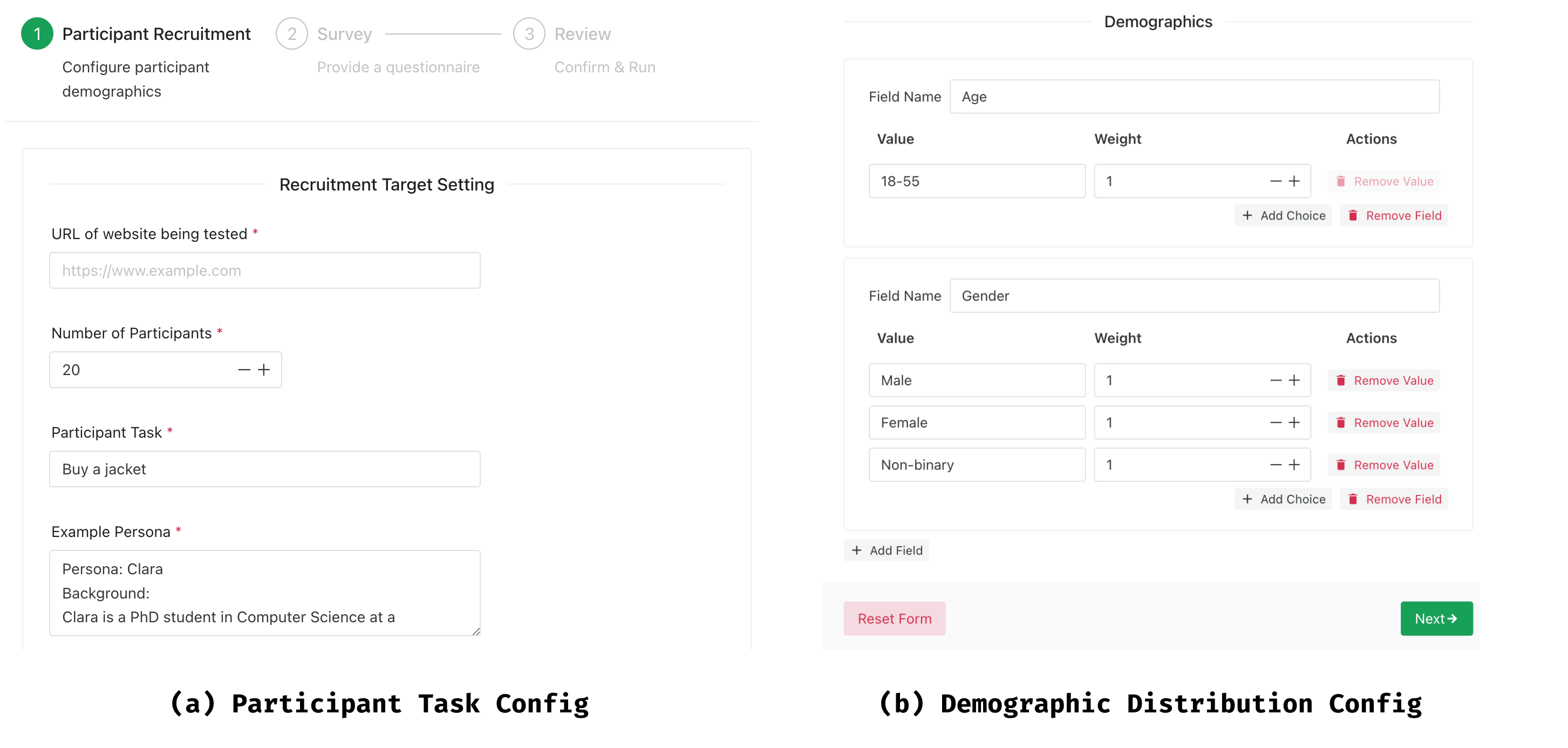}
    \caption{Study Configure Interface. Users can (a) configure the study task and provide example persona for LLM agent participants, (b) congif demographic distribution of the LLM agents. User can also configure open-ended and multiple choice quetions in a post-study survey for LLM agents to fill up after the simulated multi-turn interaction session, as show in Fig. \ref{fig:post-study-configure}}
    \Description{Screenshot of the study configuration interface with three panels.
(a) Participant Task Config — form fields to enter the website URL, number of participants, participant task, and an example persona.
(b) Demographic Distribution Config — options to add demographic fields (e.g., age, gender) with values and weights.}\label{fig:study-configure}
\end{figure*}

\subsubsection{Action Space}
\label{sec:action-space}

We design the action space so that agents can perform the same primitive interactions as human users (click, type, hover, select, navigate, manage tabs). To reduce latency and eliminate unnecessary branching, we explicitly remove the \emph{scroll} action (the environment automatically scrolls the target element into view before execution). Targets are referenced by the stable semantic identifiers described in the observation construction (\code{semantic-id}), which makes actions robust to minor DOM changes. After each action, the executor waits for network and UI quiescence (for example, no active requests for a short idle window) before returning the next observation. This design addresses prior challenges of noisy action choices, arbitrary code generation, and nondeterministic post-action states by keeping the action set compact, interpretable, and synchronized with the rendered page.

All actions that operate on elements identify their targets using semantic IDs that are guaranteed to be unique within the current observation. Before execution, the environment validates that the target remains interactable, scrolls it into view automatically (therefore, no explicit scroll action is necessary), performs the operation, and then waits for an idle period before emitting the next observation. A full list of available actions can be found in appendix \ref{sec:full-action-space}

\subsection{Study Configure Interface}  
We implemented a Study Configure Interface to streamline the process of launching experiments. As shown in Fig.~\ref{fig:study-configure}, users can first specify the participant task and the URL of the website under study, along with an example persona that serves as a template for the persona generator module (Fig.~\ref{fig:study-configure}a). They can then define demographic distributions by adding fields such as age or gender and assigning weights to control sampling (Fig.~\ref{fig:study-configure}b). Finally, users can configure the post-study survey by composing open-ended or multiple-choice questions, or by importing standard instruments such as the System Usability Scale (SUS) from the database for convenience (Fig.~\ref{fig:study-configure}c). This interface allows researchers to set up complete studies quickly without additional coding effort.

\subsection{Result Viewer Interface}

\begin{figure}
    \centering
    \includegraphics[width=0.75\linewidth]{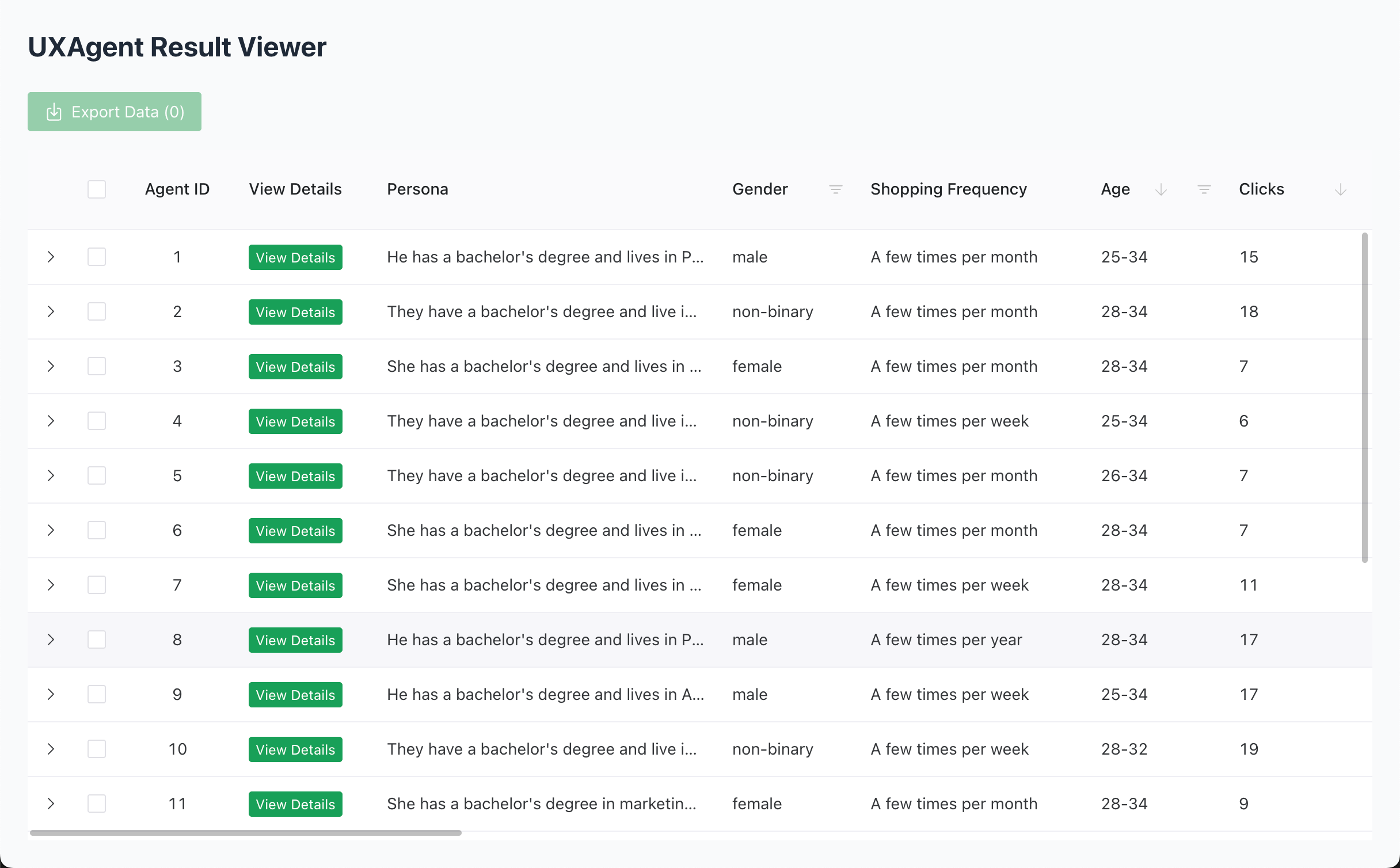}
    \caption{The home page of the Result Viewer Interface. Users can view the result table of all simulation runs and expand rows to inspect participant personas and post-study survey responses. The user can also export the data as an Excel spreadsheet.}
    \label{fig:result-viewer-homepage}
    \Description{Screenshot of the UXAgent Result Viewer interface. A table lists simulated participants with columns for Agent ID, persona, gender, shopping frequency, age, and number of clicks. Each row has a button to view details, and an export button appears above the table.}
\end{figure}

\begin{figure}
    \centering
    \includegraphics[width=.95\linewidth]{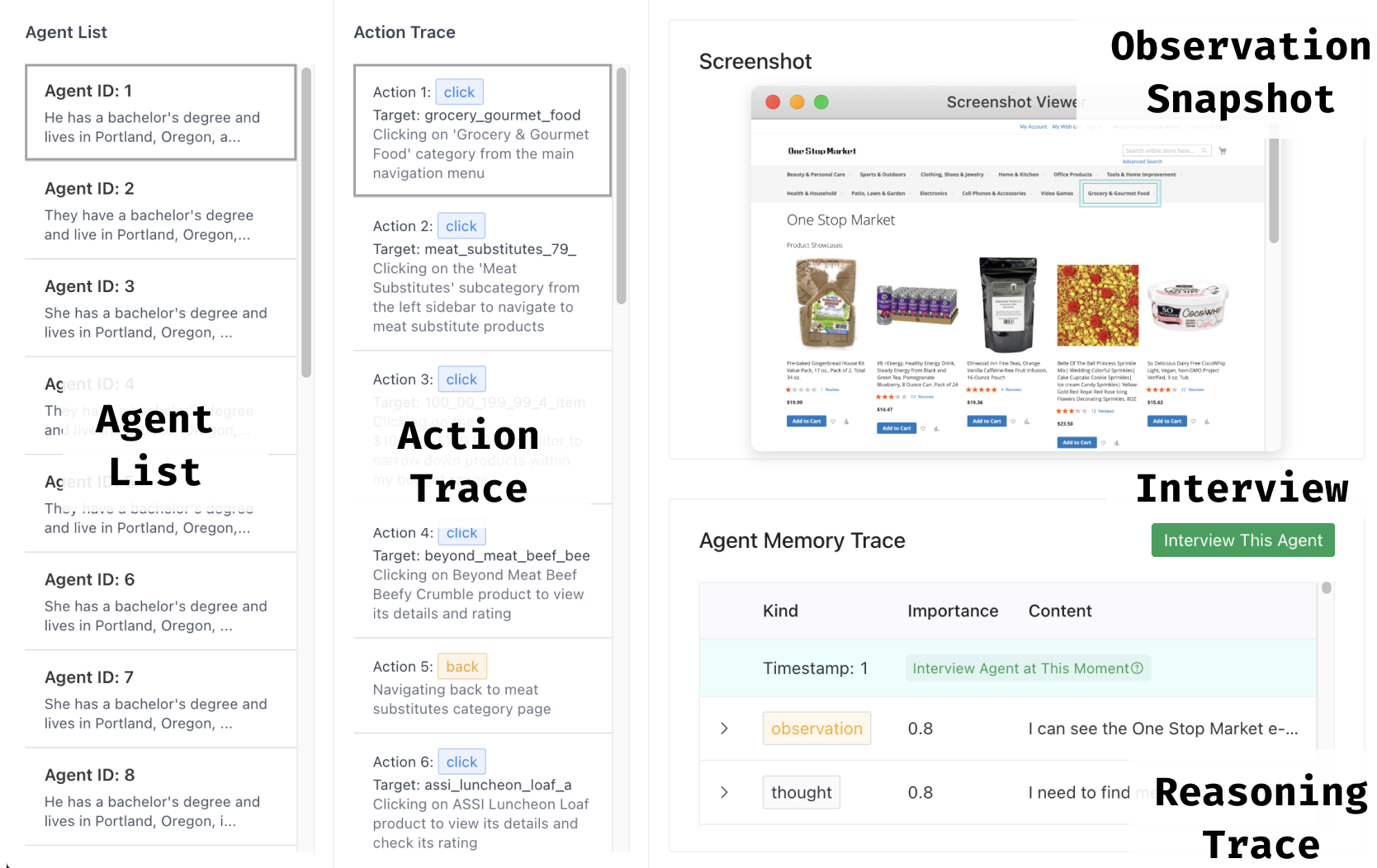}
    \caption{Detail view of the Result Viewer. A list of all agents is displayed on the left. Clicking on one of the agents will display it's action trace, reasoning trace and observation snapshot. Clicking on the interview button will show the interview interface pop-up window (Fig. \ref{fig:interview})}
    \label{fig:result-viewer-detail}
    \Description{Screenshot of the UXAgent Result Viewer detailed view. The left panel shows an agent list with IDs and persona descriptions. The middle panel displays an action trace with step-by-step clicks and navigation. On the right, there is a screenshot snapshot of the shopping website, an interview button labeled “Interview This Agent,” and a reasoning trace table showing observations and thoughts with timestamps, importance scores, and content.}
\end{figure}
The Result Viewer provides a comprehensive interface for analyzing simulation results. As shown in Fig.~\ref{fig:result-viewer-homepage}, the home page displays a table containing all simulation runs, including avent personas, demographics, quantitative measures, and post-study survey results. Users can filter and sort the table headers or export selected runs for external analysis. 

By clicking on the \textit{View Details} button, users can enter the detail view (Fig.~\ref{fig:result-viewer-detail}), which presents the agent’s interaction trace step by step. The action trace records each model action (e.g., clicks or text inputs), while the screenshot viewer displays the corresponding web page state at that moment. The target element of the current action is highlighted with a blue outline in the screenshot. In parallel, the reasoning trace viewer shows the agent's internal reasoning, including observations, thoughts, and reflections, which can be expanded for detailed inspection.  

\begin{figure*}
    \centering
    \includegraphics[width=0.5\linewidth]{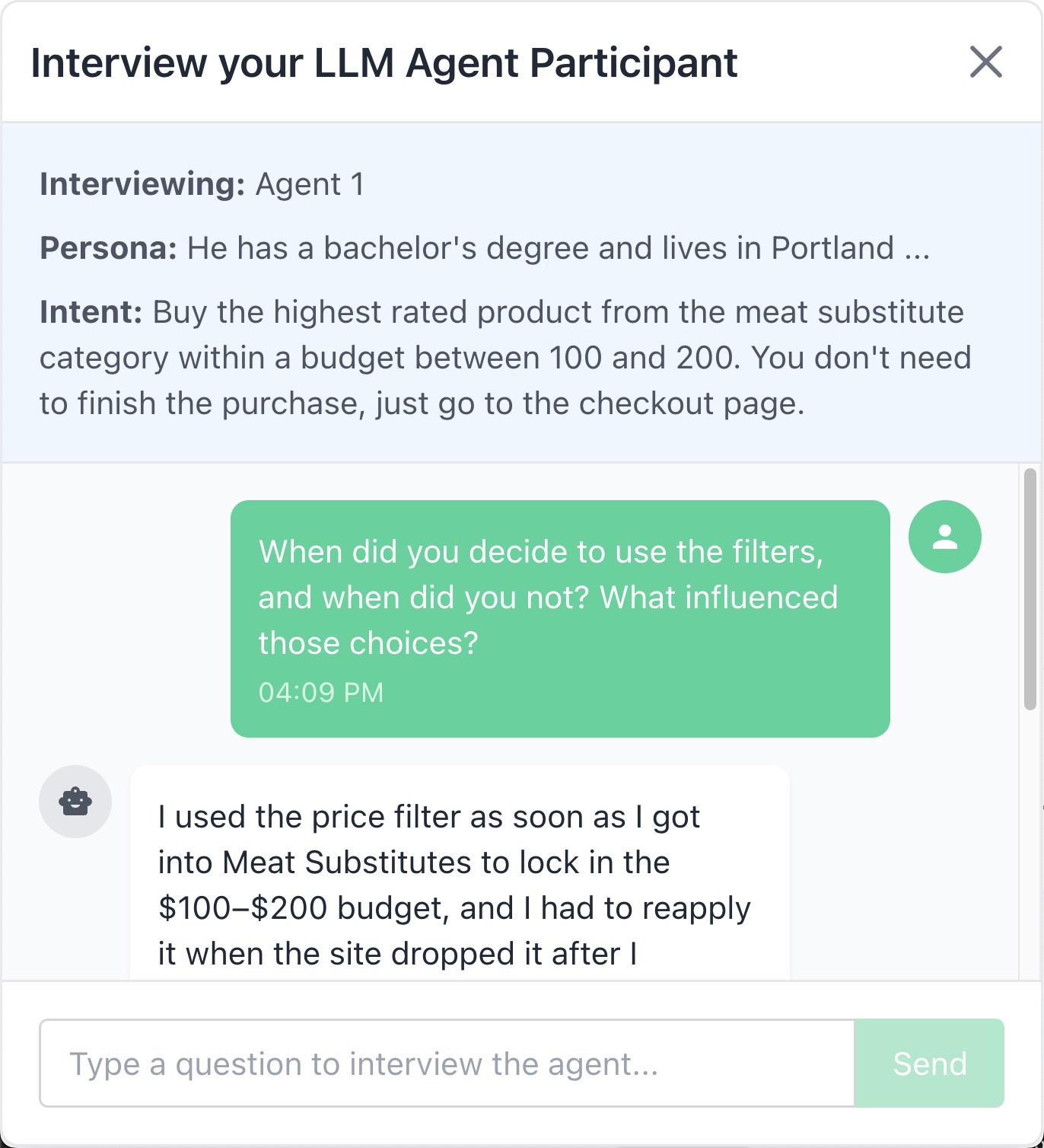}
    \caption{Interview Interface. Researchers can interview the LLM agent participants at any time point to mimic a contextual inquiry in the middle of the UX study session or a post-study interview at the end. The agent responds based on its persona, intent, and reasoning trace till that moment.}
    \Description{ Screenshot of the interview interface. Researchers can ask LLM agent participants open-ended questions either post-study or mid-study. The panel shows the agent’s persona and intent at the top, followed by a chat exchange where the researcher’s question appears in green and the agent’s detailed response appears below. A text box at the bottom allows researchers to type new questions.}\label{fig:interview}
\end{figure*}

The system also supports interactive post-study interviews (Fig.~\ref{fig:interview}).
Researchers can conduct interviews after an LLM agent participant has completed its interaction with the website, allowing them to collect qualitative feedback.
Interviews may also be injected mid-simulation at a specific timestamp.
In this case, the agent responds based only on its memory up to that point, without knowledge of future events, enabling researchers to probe reasoning in real time.
Unlike human participants, whose answers may be influenced by confirmation bias \cite{klayman1995varieties} when recalling earlier steps, LLM agents can be interviewed using a precise memory snapshot at a given moment, providing an accurate window into their decision-making process.

\section{Case Study: A Usability Testing Simulation with \projectname to Evaluate a Newly Designed Product Filter Feature in an Online Shopping Website }

We chose online shopping, a common and highly personalized use case, to illustrate how UX researchers can use \projectname to augment their usability testing work practice.
In our evaluation study, we used WebArena~\cite{zhouWebArenaRealisticWeb2024} as our target online shopping web environment.
We chose WebArena as our experiment platform because it works similarly to Amazon.com and is a publicly available online shopping environment, widely used in many web agent studies\cite{suLearnbyinteractDataCentricFramework2024, yangAgentOccamSimpleStrong2024}.

\subsection{LLM Agent's Usability Testing Task and Simulated Participant Data}

In our study, we first created an initial usability testing study protocol borrowed from prior HCI works \cite{Moser2017, prakash2024}.
We employed the Persona Generator Module to create 20 agent personas, ensuring a uniform distribution across gender (male, female, and non-binary) and online shopping frequency (a few times per year, a few times per month, and a few times per week). 
LLM agent module uses each persona to perform a usability testing task: \textit{``Buy a mean substitute. With the highest rating and within a budget between \$100 and \$200. You don't need to finish the purchase, just go to the checkout page.''}
LLM agents were free to explore the WebArena shopping environment via multi-turn interaction to test the newly designed feature of ``product filtering panel''~\cite{zhouWebArenaRealisticWeb2024}.  
They may or maynot use the filter naturally, or they could either complete a purchase or terminate the session without purchase, if they could not find a suitable item.
During each session, our system recorded the full simulation result, including the LLM agent participants' reasoning and action traces, as well as their post-study surveys.
Examples of the generated data are shown in appendix \ref{sec:example-data}.
As a reference to the system's scalability, running 20 simulations took approximately 90 minutes, and 1,000 simulations took about 12 hours.
The overall runtime primarily depends on the LLM API's rate limits or token throughput.

\begin{table*}[t]
    \centering
    \caption{Participant demographics (age, gender, years of experience with UX research (YoE), and self-reported familiarity with usability testing (1 to 5 very familiar)) and post-study survey responses on usability (-2 to +2 strongly agree), helpfulness (-2 to +2 strongly agree), trust (-2 to +2 strongly agree), satisfaction with revised study design (-2 to +2 strongly agree), and whether LLM can replace human participants (-2 to +2 strongly agree).}
    \label{tab:findings}
\begin{booktabs}{
  colspec = {ccccccccc},
  row{1} = {font=\bfseries},
  cell{1}{4-9}={font=\footnotesize\bfseries},
  cells={c,m}
}
\toprule
PID & Gender & YoE & {UX Familarity \\ (1 to 5)} & {``Easy to Use'' \\ (-2 to 2)} & {``Helpful'' \\ (-2 to 2)} & {``Trust'' \\ (-2 to 2)} & {``Satisfied with Final\\ Study Design'' (-2 to 2)} & {``LLM Can Replace \\ Humans'' (-2 to 2)} \\
\midrule
P1  & Female & 5+   & 5 & 1 & 2 & 1  & -1 & -1 \\
P2  & Male   & 1--2 & 3 & 2 & 2 & 1  &  1 & -1 \\
P3  & Female & 5+   & 5 & 1 & 1 & 1  &  1 &  0 \\
P4  & Male   & 2--3 & 3 & 1 & 1 & 0  &  0 & -2 \\
P5  & Male   & 1--2 & 3 & 1 & 1 & 0  &  1 & -1 \\
P6  & Female & 0--1 & 3 & 2 & 0 & 0  &  1 & -1 \\
P7  & Female & 5+   & 5 & 2 & 2 & 1  &  2 & -1 \\
P8  & Female & 3--4 & 4 & 2 & 1 & 0  &  0 & -2 \\
P9  & Male   & 5+   & 4 & 1 & 1 & 0  &  0 &  1 \\
P10 & Female & 4--5 & 5 & 2 & 2 & 2  &  1 &  1 \\
P11 & Female & 1--2 & 3 & 1 & 2 & 1  &  1 & -1 \\
P12 & Female & 1--2 & 3 & 2 & 0 & 0  &  2 & -1 \\
P13 & Male   & 3--4 & 4 & 2 & 2 & 2  &  2 &  2 \\
P14 & Female & 4--5 & 5 & 2 & 1 & -1 &  2 & -1 \\
P15 & Female & 1--2 & 3 & 2 & 2 & 1  &  1 &  0 \\
P16 & Female & 3--4 & 5 & 2 & 2 & 1  &  2 &  1 \\
\midrule
Average & -- & -- & -- & 1.63 & 1.38 & 0.63 & 1.00 & -0.44 \\
SD      & -- & -- & -- & 0.50 & 0.72 & 0.81 & 0.89 & 1.15 \\
\bottomrule
\end{booktabs}
\end{table*}

\subsection{User Study with 16 UX Researchers}
\label{sec:study-protocol}

We recruited 16 UX researchers to participate in a heuristic evaluation study with reviewing the 20 simulated sessions using \projectname. 
The detailed demographics and experience levels of these UX researcher participants are summarized in Table~\ref{tab:findings}. Five participants reported more than 5 years of UX research experience, two reported 4--5 years, three reported 3--4 years, one reported 2--3 years, five reported 1--2 years, and one reported less than one year. Most participants rated themselves as 3 within 1 to 5 in usability testing familiarity (M=3.87, SD=0.92).

Each study session lasted approximately 30 minutes and was video recorded if they consented. Participants were first presented with the initial draft of the usability testing study design protocol (included in Appendix \ref{sec:study-protocil-draft}); if they are unsatisfied, they are instructed to make suggestions to revise the study design. 
They then used the Study Configure Interface  to config the study design in \projectname. 
To accommodate the time limit, participants were provided with results from a pre-configured run rather than executing new simulations within the 30 mins of study sessoin in real-time. 
Using the Result Viewer Interface, they reviewed and analyzed the generated data of the 20 LLM agent participants. 
UX researcher participants also conducted a post-study interview with the LLM agent, following the interview protocol defined in the draft study design. 

\begin{figure}[t]
    \centering
    \includegraphics[width=0.8\linewidth]{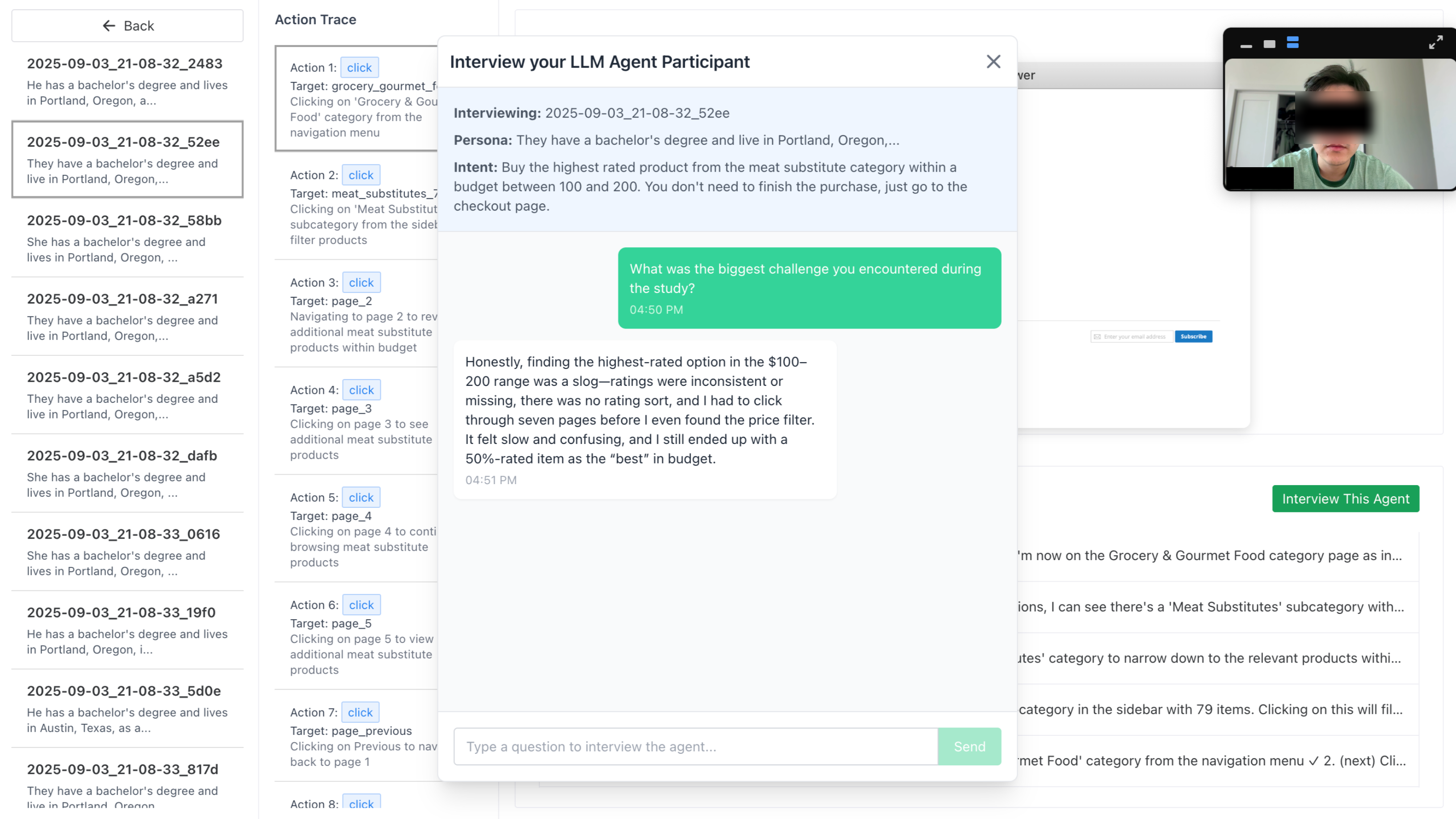}
    \caption{One of our 16 UX researcher participants uses \projectname in a session.}
    \Description{Screenshot of a participant using the UXAgent system during the study. The interface shows the agent’s action trace on the left, an interview window in the center where the researcher’s question (in green) and the agent’s response are displayed, and a video call window with the participant at the top right.}\label{fig:participant}
\end{figure}

After completing the data analysis and interviewing the agent, UX researchers were asked to propose improvement suggestions both to the initial study protocol and to the new ``filter feature'' design on the shopping website. 
Finally, they completed a post-study survey consisting of Likert-scale items evaluating \projectname's ease of use, the helpfulness and trustworthiness of the generated data, and their satisfaction with the revised protocol. 
We also administered a structured questionnaire, which included items probing how participants think \textit{``LLM agents could replace human participants in the future usability testings'',} as well as open-ended questions to gather their feedback on \projectname's desired new features, additional data modalities, potential risks, and other qualitative feedback. A photo of one user study session is shown in Fig.~\ref{fig:participant}.

\subsection{User Study Study Findings}

\subsubsection{How Did UX Researchers Use \projectname System?}

\begin{figure*}[t]
    \centering
    \begin{subfigure}[t]{0.48\linewidth}
        \centering
        \includegraphics[width=\linewidth]{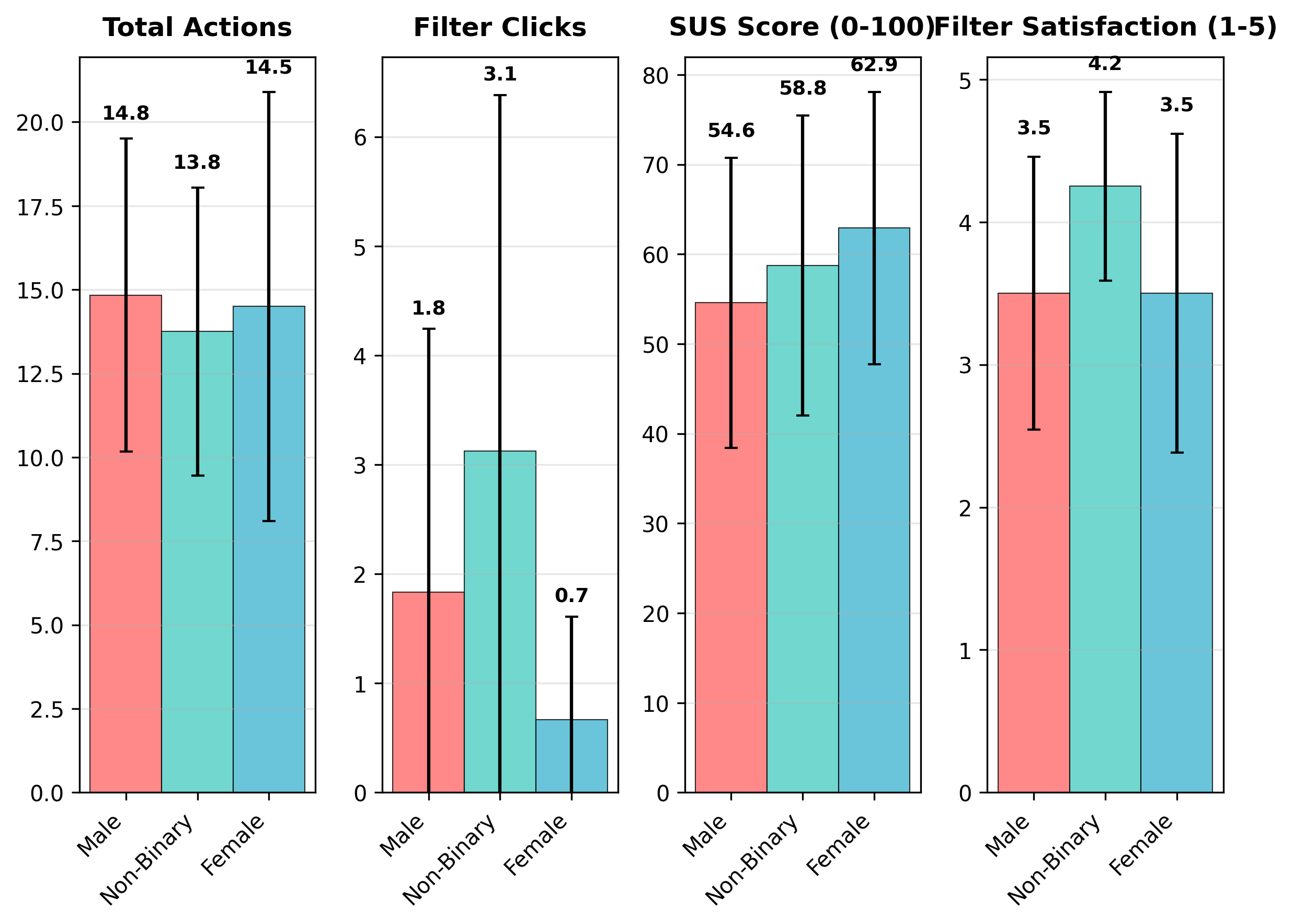}
        \caption{LLM Agent simulated online shopping behaviors, filter feature usage behaviors, and their post-study SUS score and satisfaction score of the online shopping experience, organized by persona's gender sub-groups.}
        \label{fig:chart-a}
    \end{subfigure}
    \hfill
    \begin{subfigure}[t]{0.48\linewidth}
        \centering
        \includegraphics[width=\linewidth]{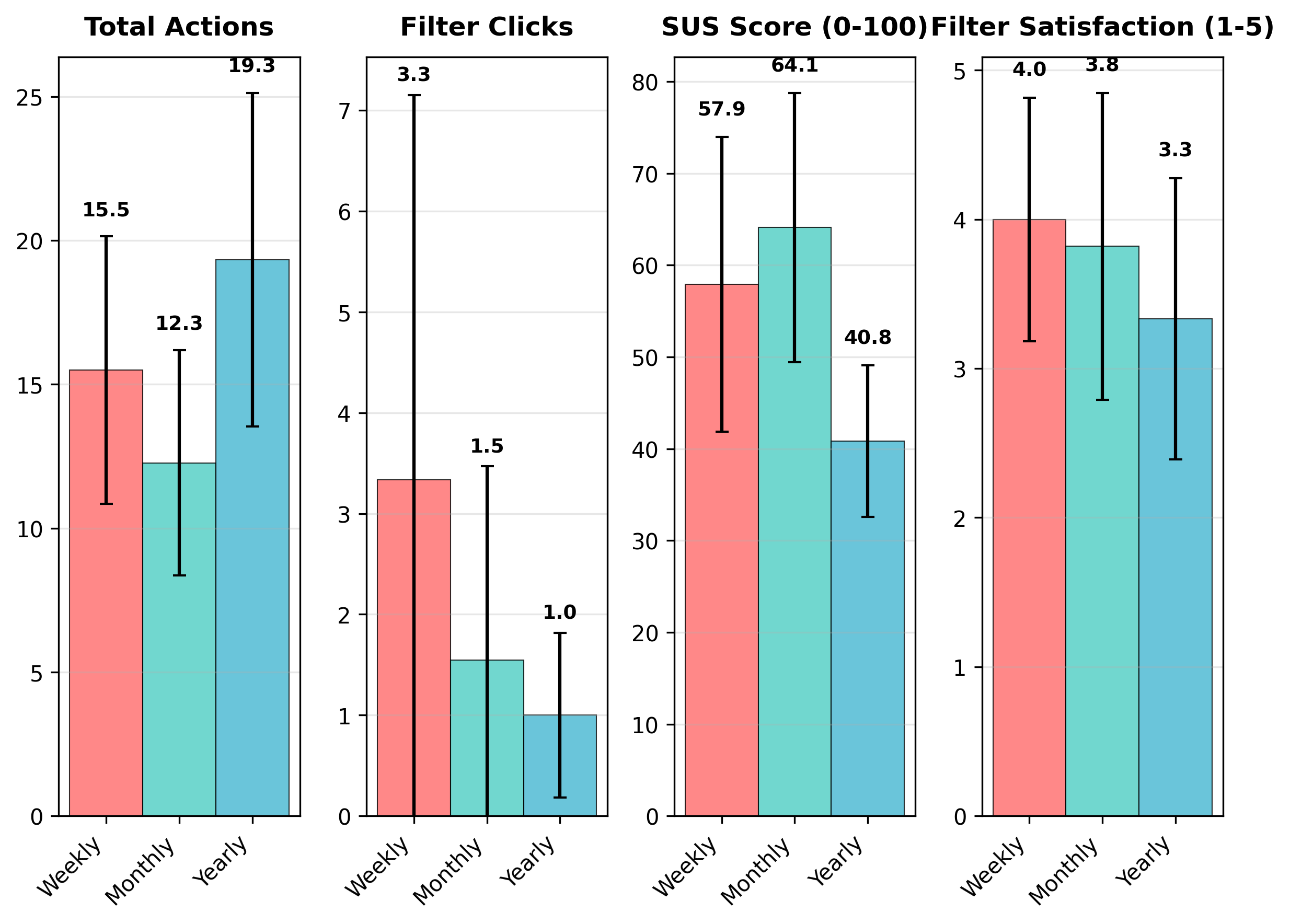}
        \caption{LLM Agent simulated online shopping behaviors, filter feature usage behaviors, and their post-study SUS score and satisfaction score of the online shopping experience, organized by persona's shopping-frequency sub-groups.}
        \label{fig:chart-b}
    \end{subfigure}
    \caption{The aggregated LLM Agent actions and their post-study questionnaire ratings on SUS and system satisfaction, illustrated by different sub-grouping strategies shown in the bar-chart visualizations.}
    \label{fig:agent-behavior}
    \Description{Left: Bar chart showing agent behavior analysis across four metrics grouped by gender. The chart contains four horizontal subplots: Total Actions, Filter Clicks, SUS Score (0-100), and Filter Satisfaction. Each subplot shows three bars representing Male (red), Female (teal), and Non-binary (blue) participants. 

In Total Actions, males show the highest average around 15 actions, females around 12 actions, and non-binary around 8 actions. For Filter Clicks, males average approximately 6 clicks, females 4 clicks, and non-binary 3 clicks. The SUS Score (System Usability Scale) shows males scoring around 50 out of 100, females around 45, and non-binary around 40. Filter Satisfaction ratings show males at approximately 4.5, females at 4.0, and non-binary at 3.5 on what appears to be a 5-point scale. All bars include error bars indicating standard deviation, and values are displayed above each bar.

Right: Bar chart analyzing agent behavior across four metrics grouped by shopping frequency. The chart displays four horizontal subplots: Total Actions, Filter Clicks, SUS Score (0-100), and Filter Satisfaction. Each subplot contains three bars representing different shopping frequencies: Weekly (red), Monthly (teal), and Yearly (blue) shoppers.

For Total Actions, weekly shoppers show the highest activity with approximately 16 actions, monthly shoppers average around 11 actions, and yearly shoppers show about 7 actions. Filter Clicks follow a similar pattern with weekly shoppers at roughly 7 clicks, monthly at 4 clicks, and yearly at 2 clicks. The SUS Score shows weekly shoppers rating the system highest at around 55 out of 100, monthly shoppers at 45, and yearly shoppers at 35. Filter Satisfaction ratings indicate weekly shoppers rate satisfaction at approximately 4.8, monthly shoppers at 4.2, and yearly shoppers at 3.0 on what appears to be a 5-point scale. All measurements include error bars showing standard deviation.
}
\end{figure*}

\paragraph{\textbf{UX researcher participants went through a four-step paradigm of using \projectname:}} 1) Building trust with data, 2) sense making of the data, 3) proposing hypotheses and in-depth exploration, and 4) drawing conclusions.

When participants had access to all types of data generated by \projectname, their predominant response was to examine each data type to ensure its trustworthiness. 
Most participants only proceeded with deeper analysis after confirming the data's trustworthiness. However, some participants began to analyze the data early on without fully verifying their trustworthiness and encountered challenges at later stages.
Once trust was established, they moved on to the sense-making phase, where they tried to interpret the data and understand the relationships between different variables.
P3 used the sorting feature to understand different agents' behavior and picked the agent that made the most clicks on the filters to interview to understand why.
Participants then made hypotheses on the experiment result and made detailed observations to verify their hypotheses.
Most participants noticed agents were looking to sort products by their reviews, but couldn't find the system feature. 
They then checked the raw action log and interviewed the agent about their feelings about this.
Participants then discussed what they found during the analysis of the simulated result.
They organized the findings and discussed how they were going to revise and iterate their study design after the simulation.

\paragraph{\textbf{UX Researchers can use the \projectname to gather early feedback from LLM agents and iterate the study design.}}

Before the study, 12 of the 16 participants suggested minor revisions to the initial usability testing protocol, while the remaining participants agreed that the original design was ready for a human pilot study.
The proposed changes included adding additional measurements, increasing the number of participants, and, in some cases, modifying the system’s UI design.
Afterward, all 16 participants were satisfied with the revised study protocol and proceeded to use \projectname.
During each session, the UX researcher participants followed our study protocol, as described in Sec.~\ref{sec:study-protocol}, to review the LLM agent–simulated usability testing data, conduct interviews, and perform analyses.
At the end of the session, each participant also completed a post-study questionnaire with open-ended questions to report their perceived trustworthiness and satisfaction with \projectname.

The result shows that although all of them were satisfied with their initial study protocol, almost all participants (14 out of 16) proposed additional improvements to the study design with the help of the \projectname.
Of the 14 participants who proposed additional improvements to the study design, 6 suggested changes to the participant tasks (P1, P2, P3, P4, P6, P14), such as adding additional filtering tasks, or ensuring task alignment with study goals.
Three participants recommended recruiting from different or more diverse demographic groups (P5, P10, P12) to avoid bias and better capture varied perspectives:
\begin{quote}[P5]
...add restrictions on the highest level of degree of participants since all of them seem to be bachelor's graduates. This might cause some bias to the study.
\end{quote}
Two participants proposed adding or refining survey questions (P7, P11, P15), and another two suggested expanding or improving the interview protocol to capture more comprehensive user experiences (P7, P8).
Additionally, two participants recommended incorporating more metrics or measurements into the study design (P13, P14).
These results show that \textbf{\projectname{} can effectively support early feedback collection for iterating on study designs}.

\paragraph{\textbf{\projectname supports UX researchers to come up with new design ideas of system features}}
Regarding the to-be-evaluated feature design of the system, participants also made use of the simulated data generated by \projectname. 
For example, seven participants explicitly suggested adding or refining a rating-based sorting or filtering mechanism (P2, P3, P5, P6, P7, P10, P11).
Others proposed additional enhancements such as customizable price filters (P2, P5) and more detailed product tags (P2). 
\begin{quote}[P2]
Add sort and filter by rating, price buckets. improve category results.
\end{quote}
Several participants also recommended improvements to the usability and flexibility of the filter panel, including multi-select options, better category design, and enhanced ranking and relevance (P12, P13, P14, P16).
\begin{quote}[P16]
Include badges/markers specific to the item being purchased in the filters. For example, for clothings, maybe the brand, the use (e.g., casual), and the colors.
\end{quote}
Finally, two participants pointed out interface issues with the filter panel and recommended making it sticky to remain visible during navigation (P4, P9), while another suggested improving its visual prominence (P14).
These findings suggest that \projectname{} can effectively support UX researchers to come up with new design ideas for the system feature, which serves the ultimate goal of the usability testing method.

\subsubsection{Participants' Perceptions and Concerns of \projectname}

\paragraph{\textbf{Participant found \projectname easy to use and helpful}}
Table~\ref{tab:findings} summarizes participants' ratings of UXAgent's usability, helpfulness, and trustworthiness on a five point (-2 to 2) Likert scale. Participants rated the system as very easy to use ($M=1.57$, $SD=0.51$), indicating that they could learn and operate the interfaces smoothly within the 30-minute session.  
They also rated the data generated by \projectname as helpful ($M=1.29$, $SD=0.73$). Participants reported that the simulated data provided meaningful insights that supported early revisions of usability testing protocols. 
Finally, participants expressed trust in the generated data ($M=0.57$, $SD=0.85$).

When asked how they would use the system in future research, most participants indicated that \projectname could help them obtain early feedback prior to conducting large-scale studies:  
\begin{quote}[P1]
I guess I will use it to generate some data before a large-scale implementation, and also help me to get some inspiration.
\end{quote}

\paragraph{\textbf{Participants have mixed feeling on whether the simulated data can represent real human participants.}} 

Some participants felt that the LLM-generated behavior ``makes sense'' and looks ``rational''.
Others explicitly noted that some behaviors appeared unrealistic:  
\begin{quote}[P6]
Some behaviors are abnormal---not what real users would do.
\end{quote}

When asked whether they trusted the generated data from \projectname, almost all participants expressed somewhat trust (M=0.63, SD=0.81), whereas P14 gave a negative rating (-1) on the five-point Likert scale.

Some participants also suggested that knowing more technical details of UXAgent, such as the fine-tuning data, would increase their trust in the results:  
\begin{quote}[P16]
...it would depend on whether I know how the UXAgent's LLM has been trained. If I know it's been trained based on data related to my field or there's some metrics proving its accuracy, I would trust it.
\end{quote}

Most interestingly, when asked whether LLM agents could replace human participants, most participants expressed disagreement ($M=-0.57$, $SD=1.16$). 
One participant explains her position by saying:
\begin{quote}[P1]
I'm really afraid that researchers only rely on this and overtrust such data.
\end{quote}

\section{Discussion}

\subsection{Design Considerations for  LLM Agent Systems in User Study}
\label{sec:discussion-design-considerations}

The shift towards using LLMs to assist UX research has demonstrated considerable potential. With \projectname, we show that LLM agents can support user studies by enabling faster design iterations through user behavior simulation, thereby reducing costs and improving the quality of study design~\cite{decker-maurerMethodMadnessUsability2012}. Building on these findings, we outline several design considerations to guide the development of future systems involving AI-simulated agents in user studies.  

\paragraph{Generate high-level insights from raw data.}  
Future systems should help researchers move beyond raw reasoning traces by automatically generating high-level insights. As P5 noted, visualizations beyond simple aggregate measures (e.g., mean values) would make it easier to intuitively identify issues. Similarly, P3 reported feeling overwhelmed by the detail page, suggesting that concise summaries of memory data would improve usability.  

\paragraph{Support more modalities of simulation and result.}  
Future systems should extend beyond text-based interaction and static outputs~\cite{liu2025personaflow, kuangCollaborationConversationalAI2023}. P3 suggested enabling LLM agents to interact with systems through voice and tone, thereby broadening applicability to domains such as conversational interfaces. P4 recommended incorporating video replays in addition to step-by-step traces to better capture the dynamics of interactions. P2 further emphasized the need to expand beyond web environments to include graphical user interfaces. Together, these suggestions highlight the importance of supporting multimodal simulation and diverse result formats to improve both ecological validity and researcher insight.  

\subsection{Potential Downstream Application for LLM Agent Systems}
In this work, we explored the use of LLM Agents to simulate human participants in the context of usability testing. However, the underlying capabilities of these agents—such as reasoning, goal-driven behavior, and interaction with web environments—open the door to a variety of downstream applications beyond UX research.

\paragraph{Help with Low-Resource UX Research}  
Although LLM agent simulations cannot yet replace human participants, they can be a valuable resource in low-resource research settings. In many educational contexts, students conducting usability studies often rely on friends or family members as participants due to limited access to target users~\cite{houde2025controlling, wang2025user}. A system like \projectname could help by providing simulated participants that approximate realistic user behaviors, allowing students to practice study design, iterate on protocols, and identify usability issues before recruiting actual participants. Similarly, independent researchers or practitioners with limited budgets could use simulated agents to pilot studies quickly and cost-effectively, reducing reliance on extensive recruitment and compensation~\cite{kapania2025simulacrum}. While these simulations are not substitutes for human data~\cite{ashkinaze2025plurals}, they can significantly lower the barrier to entry for conducting UX research and provide a practical first step in the iterative design process.

\paragraph{Digital Twin for End-User Task Automation}
LLM Agents can serve as ``digital twins'', acting on behalf of users to complete tasks in web environments~\cite{vahdati2025multi}. For example, in online shopping scenarios, a future system could allow users to specify their preferences and intents (e.g., budget, brand, or product features), and delegate the search, comparison, and purchase decisions to a personalized agent. This application would enable users to offload routine decision-making while still aligning outcomes with their goals.

\paragraph{Evaluation of Existing AI Systems}

Simulated users can also serve as proxies for evaluating existing AI systems, such as recommender systems and search engines~\cite{ma2025pub, zhao2025exploring}. Unlike traditional evaluation methods that rely on historical interaction logs or static benchmarks, LLM Agents can dynamically engage with system outputs, simulate realistic user behavior, and provide feedback. By assigning diverse personas to the agents, researchers can examine how users with different goals, preferences, or constraints interact with AI systems. This simulation-based approach offers a richer, more nuanced perspective on system performance, fairness, and user alignment across varied use cases.

\paragraph{Simulating A/B Testing with Behavioral Feedback}
LLM Agents can be deployed to simulate A/B tests for different interfaces and model variants, not only measuring which version performs better in terms of task success but also collecting rich behavioral data and simulated feedback.
This enables researchers to evaluate both the quantitative and qualitative impact of design changes before allocating real A/B testing resources, potentially reducing iteration cycles and development costs.

\subsection{Resolving Risk and Privacy Concerns}

While LLM-assisted UX study systems offer substantial benefits for user studies, they also introduce significant risks and concerns, particularly around privacy and ethical use \cite{liHumanCenteredPrivacyResearch2024, shaoPrivacyLensEvaluatingPrivacy2024}. 
A notable concern about sensitive data might arise in applying LLM Agents in user studies in fields such as healthcare \cite{perisPrivacyTimeLanguage2023}.
Our participants viewed data from LLM Agents as supplementary rather than definitive, which is consistent with prior findings that LLMs cannot fully replace human subjects in psychological experiments~\cite{cui2024can}. 
This preference reflects their ability to detect biases, limiting acceptance of LLMs as accurate representations of outcomes~\cite{schmidtSimulatingHumanHCD2024}.

Another critical issue is the risk of researchers misusing AI-generated data \cite{perisPrivacyTimeLanguage2023}. 
Some may be tempted to use this simulated data as a substitute for actual human participant data, potentially bypassing real user studies altogether. 
This could lead to flawed conclusions, as the AI-generated data, while useful for simulations, cannot fully replicate the complexity of real human behavior.
We believe that addressing these risks requires a clear understanding of the limitations of LLM-generated data and establishing rigorous ethical guidelines and robust privacy safeguards to ensure responsible and effective integration of LLM in UX research.

\subsection{Limitations and Future Works}
There are several limitations in our work. 
First, the limited number of participants in our heuristic evaluation may not fully capture the diversity of user experiences or the generalizability of the findings. 
A larger scale of rigorous user testing in the future will help us further explore the effectiveness of \projectname.
Also, our current work focuses only on the qualitative analysis of the participants' perspectives toward LLM Agent simulation, but not on quantitative evaluations of the LLM Agents' versus real humans' shopping behaviors. 
It is crucial to conduct a systematic analysis comparing LLM-generated simulations with real human participants in order to further understand how UX researchers could benefit from using such agents.

Our current results confirm the effectiveness of \projectname in web-related user research, as well as the potential of LLMs in simulating user behavior to assist rapid iteration of user research. 
In the future, this work could be expanded to encompass a broader range of user research contexts beyond web applications. 
This might include adaptation to UI/UX design in desktop, mobile, or mixed-reality environments, where LLM Agents could simulate user interactions in complex, multi-modal interfaces. 
Also, we only tested giving the agents an explicit intention of ``buy a jacket''. Experimenting with other intents, such as window shopping, can also help understand the ability and limits of LLM Agent.
Additionally, the action space of the agent is currently limited to basic actions such as ``click'' and ``type''. Supporting other types of actions, such as ``wait'', ``read'', and ``scroll'' can allow the experimenter to collect more information useful to their study.

\section{Conclusion}

In this work, we present \projectname to support researchers in conducting simulated user studies.
By providing action traces, reasoning traces, and post-study survey results, the system supports iterative refinement of both UX study designs and feature designs.
Our case study with 16 UX researcher participants suggests that \textbf{\projectname helps them identify potential flaws in study protocol } and gains early insights into how users might perceive and interact with the proposed web feature, thus improving the quality and reducing risks of the later experiments with real human participants.
Our study is the first step towards a promising future of UX researchers to use LLM agents to achieve ``human-AI collaboration'' in the field of UX research.

\bibliographystyle{ACM-Reference-Format}
\bibliography{custom,set,newRef}

\clearpage
\appendix

\section{Full Action Space}
\label{sec:full-action-space}
\begin{itemize}[leftmargin=*]

\item {\textbf{A. Element-level interactions}}
  \begin{itemize}[leftmargin=*]
    \item {Click element}: perform a click action on the target element such as a button, link, or control.  
    \item {Hover element}: move the cursor over the target element to trigger tooltips or dropdown menus.  
    \item {Key press}: send a keyboard event such as Enter, Escape, Tab, or arrow keys, optionally focusing an element first.  
  \end{itemize}

\item {\textbf{B. Form and text input}}
  \begin{itemize}[leftmargin=*]
    \item {Type text}: enter text into an input field or editable region, optionally pressing Enter afterwards.  
    \item {Clear input}: remove all content from an input field or editable region.  
    \item {Select option}: choose a specific option from a dropdown or select menu.  
  \end{itemize}

\item {\textbf{C. Navigation and page control}}
  \begin{itemize}[leftmargin=*]
    \item {Navigate to URL}: load a new address in the current tab.  
    \item {Back}: navigate one step backward in the browser history.  
    \item {Forward}: navigate one step forward in the browser history.  
    \item {Refresh}: reload the current page.  
  \end{itemize}

\item {\textbf{D. Tab management and task control}}
  \begin{itemize}[leftmargin=*]
    \item {New tab}: open a new browser tab, optionally with a specified URL.  
    \item {Switch tab}: change focus to another tab by index.  
    \item {Close tab}: close an existing browser tab.  
    \item {Terminate task}: shut down the browser session and optionally submit a final answer.
  \end{itemize}

\end{itemize}

\begin{figure}
    \centering
    \includegraphics[width=0.5\linewidth]{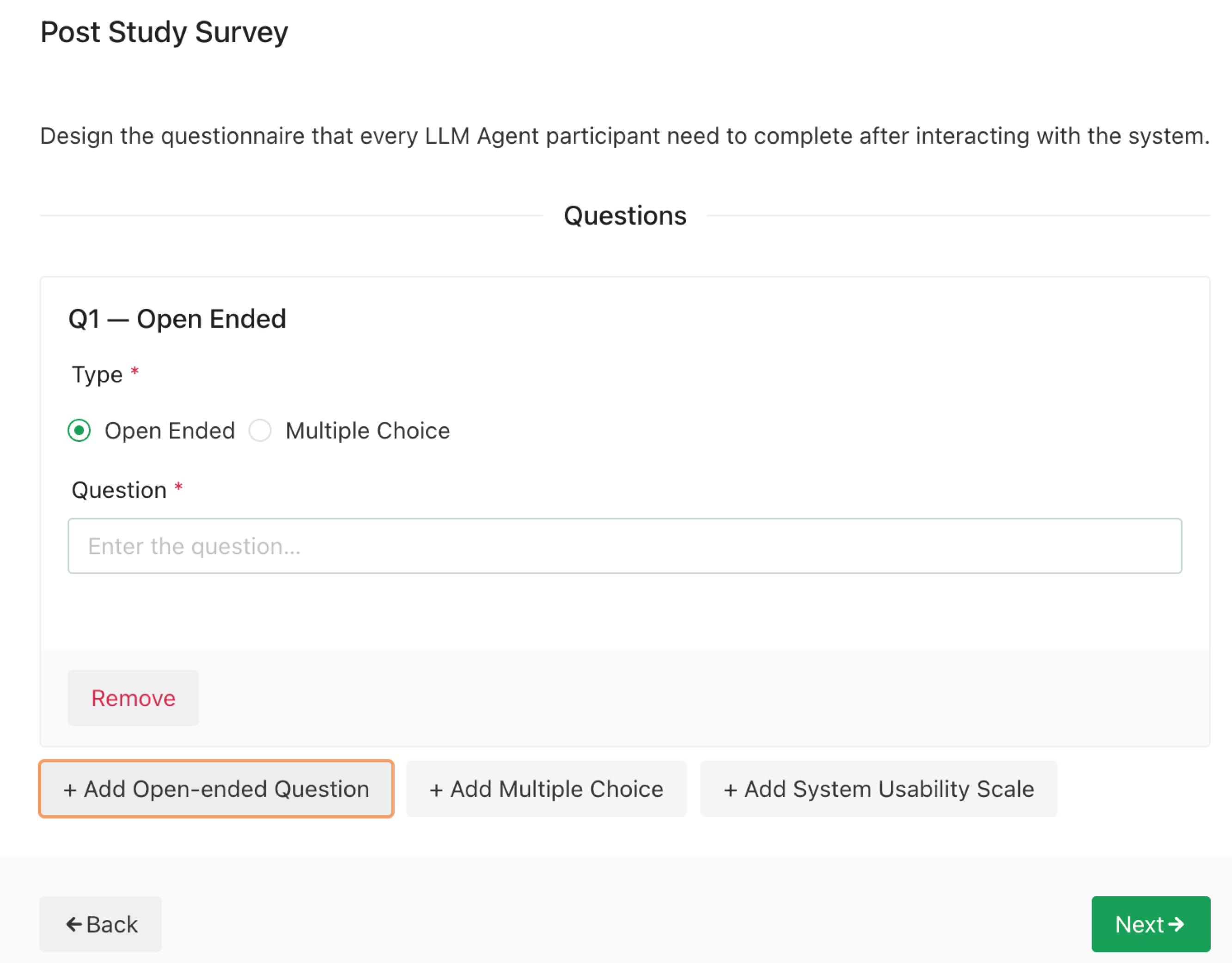}
    \caption{Post-study survey configure interface.}
    \label{fig:post-study-configure}
\end{figure}

\section{Study Protocol Draft}
\label{sec:study-protocil-draft}
\section*{Study Context}
You are the designer of a shopping website and have just completed the implementation of a new ``left filter panel'' on the search results page.

\section*{Draft Study Protocol}
The goal of this study is to evaluate the usability and effectiveness of the left filter panel in supporting users' search and purchase tasks. 

\subsection*{Study Procedure}
Participants are asked to:
\begin{enumerate}
    \item Finish the participant task on the shopping website
    \item Finish a survey questionnaire
    \item Finish a post-study interview conducted by you
\end{enumerate}

\subsection*{Measurements}
\begin{itemize}
    \item Click count: total number of clicks made by the user
    \item Filter click count: total number of clicks on the filters
    \item Step count: total number of steps taken to complete the task
\end{itemize}

\subsection*{Participant Task}
Buy the highest rated product from the meat substitute category within a budget between 100 and 200. You don't need to finish the purchase, just go to the checkout page.

\subsection*{Participants}
\begin{itemize}
    \item Total: 20 participants
    \item Age range: 18--55 
    \item Gender: Male, Female, Non-binary (1:1:1)
    \item Online Shopping Frequency: evenly distributed across three frequency groups (1:1:1)
    \begin{itemize}
        \item A few times per year
        \item A few times per month
        \item A few times per week
    \end{itemize}
\end{itemize}

\subsection*{Post-Study Survey Questions}
\textbf{Usability (1 = Strongly disagree, 5 = Strongly agree):}
\begin{enumerate}
    \item I think that I would like to use this system frequently.
    \item I found the system unnecessarily complex.
    \item I thought the system was easy to use.
    \item I think that I would need the support of a technical person to be able to use this system.
    \item I found the various functions in this system were well integrated.
    \item I thought there was too much inconsistency in this system.
    \item I would imagine that most people would learn to use this system very quickly.
    \item I found the system very cumbersome to use.
    \item I felt very confident using the system.
    \item I needed to learn a lot of things before I could get going with this system.
\end{enumerate}

To what extent do you agree that the left-side filter function helps you narrow your search and find your desired product? (1 = Strongly disagree, 5 = Strongly agree)

Please explain your answer. What aspects of the filter function were helpful or unhelpful in your experience?

\subsection*{Interview Questions}
\begin{enumerate}
    \item When did you decide to use the filters, and when did you not? What influenced those choices?
    \item What do you think is the main benefit of the new filter functionality?
    \item What challenges or downsides did you encounter when using the filters?
\end{enumerate}

\section{Prompt}
\subsection{Persona Generation}
\label{sec:prompt-persona}
\begin{minted}[breaklines, breaksymbol={},breakanywhere]{markdown}
You are a helpful assistant that generates diverse personas.
Examples: <EXAMPLE PERSONA>

Generate a persona using the above examples. The persona should be different from previous personas to ensure diversity.
The persona should:
- have the [DEMOGRAPHIC]
Provide the persona in the same format as the examples.
Only output the persona, no other text.
\end{minted}

\subsection{Perceive Module}
\begin{minted}{markdown}
You are a module within an automated web agent tasked with simulating a user's interaction with a webpage. Your specific role is the PERCEIVE module.

You will be provided with an environment that includes:
- Full HTML content of the current page.
- A list of all clickable elements on the page
- A list of all input elements on the page
- A list of all select elements on the page

Your task is to simulate how a human would perceive the entire current page. You must output a **single observation**, in natural language, describing everything visible on the page exactly as it appears. Output this as a JSON array with **only one item**.

You are not allowed to summarize or interpret. Your only job is to **translate the HTML content into natural language**, as completely and faithfully as possible. Every piece of visible text must be preserved. If the page contains product listings, dialog boxes, or menus, you must describe **all text, labels, and options** exactly as shown.

**Use the following format:**

```{"observations": ["<a single, fully detailed paragraph describing everything on the current page>"]}```

**Rules you must follow**:
- Output only one observation in the list. It should describe the **entire HTML content** provided.
- Do **not** summarize, skip, paraphrase, or omit any visible information. Every piece of text in the HTML must be captured verbatim.
- Think in the first person. Write what a user would perceive by looking at the screen.
- Do **not** mention or refer to the structure of the HTML itself.
- The provided lists of buttons and inputs are for reference only -- only include them if they appear in the actual HTML content.
- Do not use any ordering words like "first", "at the top", or "below".
- If the HTML is empty, return an empty list: `"observations": []`.

**DO NOT OUTPUT ANYTHING ELSE. DO NOT OUTPUT ANYTHING ELSE. DO NOT OUTPUT ANYTHING ELSE. DO NOT OUTPUT ANYTHING ELSE. DO NOT OUTPUT ANYTHING ELSE.**

<IMPORTANT>
INCLUDE EVERY DETAIL ON THE WEB PAGE, as much as possible. at least include all the text on the page. All input text should be included in your output.
INCLUDE EVERY DETAIL ON THE WEB PAGE, as much as possible. at least include all the text on the page. All input text should be included in your output.
INCLUDE EVERY DETAIL ON THE WEB PAGE, as much as possible. at least include all the text on the page. All input text should be included in your output.
INCLUDE EVERY DETAIL ON THE WEB PAGE, as much as possible. at least include all the text on the page. All input text should be included in your output.
INCLUDE EVERY DETAIL ON THE WEB PAGE, as much as possible. at least include all the text on the page. All input text should be included in your output.
INCLUDE EVERY DETAIL ON THE WEB PAGE, as much as possible. at least include all the text on the page. All input text should be included in your output.
INCLUDE EVERY DETAIL ON THE WEB PAGE, as much as possible. at least include all the text on the page. All input text should be included in your output.
INCLUDE EVERY DETAIL ON THE WEB PAGE, as much as possible. at least include all the text on the page. All input text should be included in your output.
INCLUDE EVERY DETAIL ON THE WEB PAGE, as much as possible. at least include all the text on the page. All input text should be included in your output.
</IMPORTANT>
\end{minted}
\subsection{Planning Module}
\begin{minted}{markdown}
# General Guidelines
You are tasked with creating a detailed plan based on the given persona, intent, and relevant memories (including observations, reflections, and previous actions if any).

Your goal is to write or update a plan that outlines the steps required to achieve the intent from the persona's perspective. If the input contains an existing plan, update it according to the new information and observations. If there is no existing plan, create a new one from scratch.

The previous plan, if any, is from the last step.

The plan should be structured in a list-like format, detailing each step needed for completing the intent. Always adhere to the preferences and characteristics of the persona (e.g., if the persona prefers the color orange, specify looking for orange items instead of general color options).
Each step will be translated to a series of browser actions. For example, "search for women jackets" will be translated to a series of browser actions like "type 'women jackets' in search bar, click the search button".
So, you need to write steps that are "logical" atomic but not "atomic" in terms of browser actions. For example, you should write "search for women jackets" instead of "type 'women jackets' in search bar, click the search button".
Each step should represent a logical action. Example:
* Search for jackets
* Filter results by preferred color (red)
* View product details for promising options
* Add selected jacket to cart
* Proceed to checkout and purchase the jacket.
For **FUTURE STEPS**, you can write if statements like "If the product matches my preference, add it to cart, else continue browsing other products".
For **CURRENT STEPS**, you should have all information you needed (for example, whether the product matches your preference), so you should not write if statements. You should write concrete, actionable steps.
The plan should be concrete. For example, use "view product details for 'Women's Winter Puffer Jacket - Black'" instead of "browse products". If any required information is missing, add a placeholder like "view product with $product_name from search results".
You should think in the first person.

**Clarification on Steps**
Avoid breaking down steps into individual browser actions (e.g., typing in the search bar, clicking buttons). Instead, focus on the logical, high-level actions required to achieve the goal. For example:
- Write "Search for jackets" instead of "type 'jackets' in search bar and click search button"
- Write "Filter by color" instead of "click the color filter dropdown and select red"
- Write "Add to cart" instead of specifying the process of clicking the add_to_cart button.

<important>
**Rationale**
DO NOT MENTION THE RATIONALE IN THE STEPS. You should say "Search for winter jackets", not "Search for winter jackets to find options that match my style preferences".
ALWAYS PUT THE RATIONALE IN THE RATIONALE FIELD. DO NOT MENTION THE RATIONALE IN THE STEPS.
</important>

Your goal is to describe logical tasks, not the low-level mechanics of how they are performed on the website.

# Detailed Guidelines
Think in four steps:
1. **Validate Available Actions**: Before writing or updating the plan, carefully examine the current page observations to identify what actions are actually available. Check for specific UI elements, buttons, filters, and options that exist on the page. Do not plan actions that are not supported by the current interface. For example, if there are no sorting options by customer reviews visible on the search results page, do not include "sort by customer reviews" in your plan.
2. Write or update the plan. Do not rewrite any plan before the (next) step, but you can rewrite the plan after the (next) step. Revise the plan and make necessary changes based on the new information and observations AND the available actions identified in step 1. Use placeholders like $product_name for the product name. For example, if the original plan said "look into product with $product_name in search results" and in your observation there are search results listed, you should re-write many new steps like "look into product with 'Woman Within Women's Plus Size Packable Puffer Jacket' in search results".
3. Expand plan from "repeat the above steps until i find the product i want." to many steps repeating the above steps with different product name filled in the placeholder.
4. Explain the rationale behind your current next step. Think why do you make the current movement.
5. Determine which step or sub-step is (next). Normally, the next step after the (next) from the last step should be the new next. You should always move the (next) marker. You should always move the (next) marker. You should always move the (next) marker. You should always move the (next) marker. Ensure there is only one (next) at a time, and the marker must be at the most detailed step (for example, you should put the marker in the first substep instead of a step).

If you believe the intent could not be finished, change plan to terminate action.
Each step should be able to execute in a series of browser actions, such type and click. **observation is not an action**.

You will also be given a current timestamp. All memories with the same timestamp are current. Older but relevant memories will be provided for context.

Format your output as follows:

{
    "rationale": "<string representation of the plan>",
    "plan": "<string representation of the plan>",
    "next_step": "<string representation of the next step>"
}

Remember to only write valid JSON, for example, denote new lines with `\n` in strings.

# Example output
## Example 1
Initial plan (at timestep 0) for intent ("Buy a jacket"):

{
    "rationale": "I need to start by searching for jackets to find options that match my preferences.",
    "plan": "1. (next) Search for jackets in the search bar\n2. Browse the search results and filter if needed\n3. View product details for jackets that look promising\n4. Select a jacket that meets my preferences and budget\n5. Add the selected jacket to cart\n6. Proceed to checkout\n7. Complete the purchase",
    "next_step": "Search for jackets in the search bar"
}

## Example 2
After searching for jackets, the plan is updated based on available search results:

{
    "rationale": "I can see several jacket options in the search results. I need to view the details of promising products to make a decision.",
    "plan": "1. (next) View product details for 'Women's Winter Puffer Jacket - Black'\n2. Check if this jacket meets my size and style preferences\n3. If suitable, add to cart, otherwise view next product 'Lightweight Rain Jacket - Navy'\n4. Continue viewing products until I find one that matches my needs\n5. Add the selected jacket to cart\n6. Proceed to checkout\n7. Complete the purchase",
    "next_step": "View product details for 'Women's Winter Puffer Jacket - Black'"
}

# TIPS:
You should always follow these tips to write the plan.
You should always follow these tips to write the plan.
You should always follow these tips to write the plan.
You should always follow these tips to write the plan.

## ALWAYS validate what actions are available on the current page before planning steps.
## ONLY plan actions that can actually be performed based on the visible UI elements and options.
## If a desired action (like sorting by customer reviews) is not available, adapt your plan accordingly.

## Step would be actions on the web page, including typing, clicking, etc. Checking, Reading, Reviewing etc are not steps.
## Step would be actions on the web page, including typing, clicking, etc. Checking, Reading, Reviewing etc are not steps.
## Step would be actions on the web page, including typing, clicking, etc. Checking, Reading, Reviewing etc are not steps.
## Step would be actions on the web page, including typing, clicking, etc. Checking, Reading, Reviewing etc are not steps.
## Step would be actions on the web page, including typing, clicking, etc. Checking, Reading, Reviewing etc are not steps.
## Step would be actions on the web page, including typing, clicking, etc. Checking, Reading, Reviewing etc are not steps.
\end{minted}
\subsection{Action Module}
\begin{minted}{markdown}
You are operating a web browser. You will be given the current HTML and your task is to translate the next step (of natural language string format) into an JSON actionable step defined in the action space.

You are tasked with determining the **next action** based on the provided `persona`, `intent`, `plan`, `next_step`, and `environment`. Your job is to interpret `next_step` and translate it into **one** actionable step within the web environment.

All output must be a single JSON object with a top-level `actions` key whose value is an array containing **exactly one** action object that follows the **new action space** below.

---

# Input Explanation

1. **Persona**: Who I am; can influence decisions. Example: "I'm a PhD student with a stipend of $50k."
2. **Intent**: What I aim to achieve. Example: "I want to buy a jacket."
3. **Plan**: A sequence of steps that leads to the intent.
4. **Next Step**: The specific step to execute now. Example: "Select the desired size 'Small'."
5. **Environment**: A structured snapshot that includes:
   - The current page HTML (string)
   - A list of input fields by their `data-semantic-id` attributes
   - A list of clickable elements by their `data-semantic-id` attributes

Use the `environment` lists to map actions to element **targets**.

---

# Key Guidelines

1. **Think in the first person.** I perform actions step-by-step from persona, intent, plan, and environment.
2. **Translate `next_step` into a concrete UI action.** Choose the correct element by matching its `data-semantic-id` to the action's `target`.
3. **Return exactly one action** in `{ "actions": [ { ... } ] }`. The array **must contain one and only one** action object. No extra text.
4. **Use `terminate` only if the step explicitly says to stop** (e.g., "terminate", "quit", "stop"). Otherwise produce an actionable step.
5. **Do not repeat a previous action.** Even if the prior attempt failed, never output the same action again.
6. **Compress compound interactions into a single action when applicable.** For example, typing a query and submitting it is a single `type` action with `"enter": true`.
7. **Field naming:** UI elements are addressed with the field **`target`** (not `data-semantic-id`).
8. If multiple candidate targets exist, prefer an **exact match**; otherwise choose the closest normalized match (case-insensitive; ignore spaces/underscores/punctuation). **Never invent targets** not present in the environment.

---

# Output Format

Return **only** a single valid JSON object:

{
  "actions": [ { /* exactly one action object */ } ]
}

No extra text, no markdown, no commentary.

---

# Action Space

Each action object must include an `action` key specifying one of the following types. Include required fields exactly as shown.

## Click
{ "action": "click", "target": "<element_semantic_id>", "description": "Clicking ..." }

## Type (with optional Enter submit)
{ "action": "type", "target": "<input_semantic_id>", "text": "<text>", "enter": true, "description": "Typing and submitting ..." }

## Hover
{ "action": "hover", "target": "<element_semantic_id>", "description": "Hovering ..." }

## Select (e.g., dropdowns)
{ "action": "select", "target": "<select_semantic_id>", "value": "<option_value>", "description": "Selecting ..." }

## Clear (clear an input field)
{ "action": "clear", "target": "<input_semantic_id>", "description": "Clearing ..." }

## Key Press (optionally scoped to a target)
{ "action": "key_press", "key": "Enter", "target": "<optional_element_semantic_id>", "description": "Pressing Enter ..." }

## Navigation
{ "action": "goto_url", "url": "https://example.com", "description": "Navigating ..." }
{ "action": "back",     "description": "Going back ..." }
{ "action": "forward",  "description": "Going forward ..." }
{ "action": "refresh",  "description": "Refreshing ..." }

## Tabs
{ "action": "new_tab",    "url": "https://example.com", "description": "Opening in new tab ..." }
{ "action": "switch_tab", "tab_id": 1, "description": "Switching tab ..." }
{ "action": "close_tab",  "tab_id": 1, "description": "Closing tab ..." }

## Terminate (only if explicitly instructed by the step)
{ "action": "terminate", "description": "Terminating ..." }

---

# Element Targeting Rules

- Match `target` to an element `data-semantic-id` from the environment's input or clickable lists.
- You can *only* target input elements with the `type` action.
- If multiple candidates exist, prefer exact match; otherwise choose the closest normalized match (case-insensitive, spaces/underscores/punctuation treated equivalently).
- Never invent targets not present in the environment snapshot.

---

# Critical Anti-Repeat Rules

- **No matter whether the previous action was successful or not, DO NOT REPEAT A PREVIOUS ACTION.**
- **Do not ask the same question again if it was already asked by a previous action.**
\end{minted}
\subsection{Reflection Module}
\begin{minted}{markdown}
You are a human viewing a web page. You looked at the page and made a list of observations.
You will be given a list of recent memories, including observations, actions, plans, and reflection.
What are you thinking right now? Output the high-level insights you gained from the recent memories.

You will also be given a persona. You need to precisely simulate the persona. Try propose questions that this persona may ask.
You should always think in the first person.

Examples of insights:
1. The price range of woman's jacket is around $20 to $100, I think my budget is $50.
2. The product "Woman Fur Casual Vests Female Autumn Spring Fur Collar Waistcoat Ladies Sleeveless Fake Fur Coats Vest Jackets - xingse,S" is not what I want.
3. The product "Woman Fur Casual Vests Female Autumn Spring Fur Collar Waistcoat Ladies Sleeveless Fake Fur Coats Vest Jackets - xingse,S"'s review looks good to me.
4. I want to buy a product that has a high rating.
5. The products recommended by Rufus is too expensive for me.

Your output should be a JSON list of insights formatted as follows:

{
    "insights": [
        "<insight 1>",
        "<insight 2>",
        ...
    ]
}
\end{minted}
\subsection{Wonder Module}
\begin{minted}{markdown}
You are tasked with entering a stage called 'wondering,'. You are asked to 'wonder' on behalf of the given persona.

You will be provided with:
- A persona
- A memory stream

Your goal is to think as a human, generating random thoughts that the persona might have at that moment. These thoughts can be related to the context of the memory stream or entirely unrelated (such as pondering what to eat for dinner).
You should think in the first person.
You should output the 0 to 3 thoughts.

Output a series of random thoughts as they would naturally occur in a human mind.

Format your output as follows JSON:

{
    "thoughts": [
        "<random thought 1>",
        "<random thought 2>",
        ...
    ]
}

Feel free to explore a wide range of topics and ideas that might occur to the persona in the moment.
\end{minted}
\subsection{Memory Importance}
\begin{minted}{markdown}
You are tasked with evaluating the importance of a given memory in helping to achieve a specified intent from the perspective of a provided persona and the current plan (which may be empty).
Think in first person.

You will be given:
- A memory
- A persona
- An intent
- A current plan (may be empty)

Your goal is to assess how crucial the memory is for fulfilling the intent from the persona's point of view, considering the current plan.

Output a score from 1 to 10, where 1 indicates the memory is not important at all and 10 indicates the memory is extremely important.

Format your output as a JSON:

{
    "rationale": <rationale for the score>,
    "score": <score from 1 to 10>
}
### Examples of Memories
1. Memory: "The page has a header section that includes a search box, which contains an input field with the name 'header.search_box.search_input' that is currently empty and has no placeholder text."
   Output:
   {
       "rationale": "This memory is extremely importance because I want to find a product and I will use the search box to find it",
       "score": 9
   }
2. Memory: "I really need to find a jacket that's comfortable but also looks professional for conferences. Maybe something in red to keep that energy up?"
   Output:
   {
       "rationale": "This memory is not important, as it only repeats information that I already know, for example, my preferences.",
       "score": 3
   }
3. Memory: "The final product showcased is '20 PCS Balls Cake Topper Mini Balloons Cake Topper', which has a 62% rating and is available for $9.49 with an 'Add to Cart' button."
   Output:
   {
       "rationale": "This memory is not important for the intent of finding a jacket, as it relates to an unrelated product that does not aid in achieving the persona's goal.",
       "score": 1
   }
\end{minted}
\subsection{Post-study Survey}
\begin{minted}{markdown}
Purpose

I answer a structured questionnaire strictly in JSON by using: 1) the
provided questionnaire (JSON schema below), and
2) my own memories (each memory is a short text with timestamp and
optional metadata).

My job is to: - Use my memories holistically to infer the best possible
answer for each question.
- Prefer newer over older memories when they conflict, but only if both
are credible.
- Always fill every question with the most reasonable answer; never
leave blank or null.
- Never invent options that aren't in the survey.
- My output must validate against the output schema below.

Inputs

-   questionnaire: JSON object with:

    -   questionnaire_id: string
    -   title: string
    -   questions: Question[]

-   memories: array of my memory objects:

        {
          "timestamp": <the step at which the memory is generated>,
          "kind": <can be thought/action/observation/reflection>,
          "importance": <how relevant this memory was to my experience>,
          "content": <The description of the piece of memory>
        }

-   Question object:

        {
          "id": "q1",
          "type": "multiple_choice" | "short_text",
          "prompt": "string",
          "options": ["..."]  // present only for multiple_choice
        }

How I use my memories

1.  Retrieve: I identify memories whose content semantically matches the
    question's topic.
2.  Rank: I sort them by recency (timestamp descending), then by
    specificity.
3.  Resolve conflicts:
    -   If my memories disagree, I choose the most recent credible one.
    -   If still tied, I choose the most specific one.
4.  Answer every question:
    -   For multiple_choice: I select the closest option. If no direct
        match, I use "Other" if available.
    -   For text: I paraphrase concisely from my memory.
5.  Be literal but holistic: I infer answers from my overall memory
    context, not individual IDs.

Answering rules by type

-   multiple_choice:
    -   value must be one of the provided options.
-   short_text / long_text:
    -   I prefer concise paraphrase; max ~160 chars.
    -   If nothing relevant exists, I use "N/A".

Output schema (I MUST follow this exactly)

I return a single JSON object:

    {
      "questionnaire_id": "string",
      "answers": [
        {
          "question_id": "string",
          "value": "string | number | boolean",
          "reason": "verbose rationale showing memory context -> reasoning -> answer"
        }
      ],
      "meta": {
        "confidence_overall": 0.0
      }
    }

Field requirements

-   answers: one entry per question in the questionnaire (preserve
    order).
-   value:
    -   If type = multiple_choice: must be exactly one of the provided
        options.
    -   If type = text: a string (or "N/A").
-   reason: one verbose explanation (memory context -> reasoning ->
    answer).
-   meta.confidence_overall: float in [0,1], reflecting my aggregate
    certainty.

Validation & safety checks

-   I ensure every question has an answers entry.
-   For multiple_choice, I confirm value is in options.
-   No fields outside the schema. No prose outside JSON.

Example (few-shot)

Questionnaire (abridged):

    {
      "questionnaire_id": "web_shopping_experience_v1",
      "title": "Recent Web Shopping Experience Survey",
      "questions": [
        {"id":"q1","type":"multiple_choice","prompt":"Which website did you shop on most recently?","options":["Amazon","eBay","Etsy","Other"]},
        {"id":"q2","type":"multiple_choice","prompt":"How satisfied were you with the shopping experience?","options":["Very satisfied","Somewhat satisfied","Neutral","Somewhat dissatisfied","Very dissatisfied"]},
        {"id":"q3","type":"short_text","prompt":"What item did you purchase?"},
        {"id":"q4","type":"multiple_choice","prompt":"How likely are you to recommend this website to a friend?","options":["Very likely","Somewhat likely","Neutral","Somewhat unlikely","Very unlikely"]},
        {"id":"q5","type":"short_text","prompt":"What could have improved your experience?"}
      ]
    }

Memories (abridged):

    [
      {"timestamp":"2025-08-20T02:14:00Z","content":"I bought a USB-C hub on Amazon last night; checkout was smooth."},
      {"timestamp":"2025-08-06T15:10:00Z","content":"Two weeks ago, I had a rough shipping delay on eBay."},
      {"timestamp":"2025-08-20T02:20:00Z","content":"Amazon's recommendation carousel was helpful; I would shop there again."}
    ]

Expected Output:

    {
      "questionnaire_id": "web_shopping_experience_v1",
      "answers": [
        {"question_id":"q1","value":"Amazon","reason":"My most recent memory shows I purchased on Amazon, making it the latest and clearest site I used."},
        {"question_id":"q2","value":"Very satisfied","reason":"I described Amazon checkout as smooth and recommendations helpful, implying high satisfaction."},
        {"question_id":"q3","value":"USB-C hub","reason":"I explicitly mentioned the item I bought."},
        {"question_id":"q4","value":"Very likely","reason":"I stated I would shop there again, mapping to high recommendation likelihood."},
        {"question_id":"q5","value":"N/A","reason":"None of my memories indicated dissatisfaction or needed improvements."}
      ],
      "meta": {
        "confidence_overall": 0.86
      }
    }
\end{minted}

\section{Example Data}
\label{sec:example-data}

\begin{table*}[t]
\centering
\caption{Agent Behavior Data}
\label{tab:agent_behavior}
\begin{booktabs}{
colspec={ccccccc}
}
\toprule
Agent ID & Gender & Shopping Freq. & Total Actions & Filter Clicks & SUS Score & Filter Satisfaction \\
\midrule
1 & Male & Monthly & 18 & 7 & 45.0 & 2 \\
2 & Non-Binary & Monthly & 19 & 2 & 62.5 & 5 \\
3 & Female & Monthly & 8 & 2 & 75.0 & 5 \\
4 & Non-Binary & Weekly & 7 & 2 & 75.0 & 5 \\
5 & Non-Binary & Monthly & 8 & 2 & 75.0 & 5 \\
6 & Female & Monthly & 9 & 0 & 70.0 & 3 \\
7 & Female & Weekly & 16 & 2 & 72.5 & 5 \\
8 & Male & Yearly & 18 & 1 & 35.0 & 4 \\
9 & Male & Weekly & 21 & 0 & 42.5 & 3 \\
10 & Non-Binary & Weekly & 20 & 11 & 72.5 & 4 \\
11 & Female & Monthly & 11 & 0 & 75.0 & 3 \\
12 & Female & Yearly & 27 & 0 & 35.0 & 2 \\
13 & Non-Binary & Monthly & 14 & 0 & 70.0 & 3 \\
14 & Non-Binary & Monthly & 15 & 1 & 32.5 & 4 \\
15 & Female & Weekly & 16 & 0 & 50.0 & 3 \\
16 & Non-Binary & Monthly & 14 & 2 & 47.5 & 4 \\
17 & Male & Yearly & 13 & 2 & 52.5 & 4 \\
18 & Male & Monthly & 7 & 0 & 75.0 & 3 \\
19 & Non-Binary & Weekly & 13 & 5 & 35.0 & 4 \\
20 & Male & Monthly & 12 & 1 & 77.5 & 5 \\
\bottomrule
\end{booktabs}
\end{table*}

\subsection{Persona}
\label{sec:example-persona}
\begin{minted}[breaklines, breaksymbol={},breakanywhere]{text}
Persona: Michael

Background:
Michael is a mid-career professional working as a marketing manager at a technology startup in San Francisco. He is passionate about using data-driven strategies to drive growth and innovation for the company.

Demographics:
Age: 42
Gender: Male
Education: Bachelor's degree in Business Administration
Profession: Marketing Manager
Income: $75,000

Financial Situation:
Michael has a comfortable income that allows him to maintain a decent standard of living in the expensive San Francisco Bay Area. He is financially responsible, saving a portion of his earnings for retirement and emergencies, while also enjoying occasional leisure activities and travel.

Shopping Habits:
Michael prefers to shop online for convenience, but he also enjoys the occasional trip to the mall or specialty stores to browse for new products. He tends to research items thoroughly before making a purchase, looking for quality, functionality, and value. Michael values efficiency and is not influenced by trends or impulse buys.

Professional Life:
As a marketing manager, Michael is responsible for developing and implementing marketing strategies to promote the startup's products and services. He collaborates closely with the product, sales, and design teams to ensure a cohesive brand experience. Michael is always looking for ways to optimize marketing campaigns and stay ahead of industry trends.

Personal Style:
Michael has a casual, yet professional style. He often wears button-down shirts, chinos, and leather shoes to the office. On weekends, he enjoys wearing comfortable, sporty attire for outdoor activities like hiking or cycling. Michael tends to gravitate towards neutral colors and classic, versatile pieces that can be mixed and matched.

Intent:
buy a large, inflatable spider decoration for halloween

\end{minted}
\subsection{Action Trace}
\begin{minted}[breaklines, breaksymbol={},breakanywhere]{json}
{"action": "click", "target": "grocery_gourmet_food", "description": "Clicking on 'Grocery & Gourmet Food' category from the navigation menu"}
{"action": "click", "target": "meat_substitutes_79_", "description": "Clicking on 'Meat Substitutes' category from the sidebar to filter products"}
{"action": "click", "target": "100_00_199_99_4_item", "description": "Clicking on the price filter for products between $100.00-$199.99 to narrow down to items within my budget range"}
{"action": "click", "target": "add_to_cart2", "description": "Adding 'Beyond Meat Beef Beefy Crumble' to cart as it's the highest rated product within budget"}
{"action": "click", "target": "my_cart_1_1_items", "description": "Clicking on 'My Cart' to proceed to checkout page"}
{"action": "click", "target": "proceed_to_checkout", "description": "Clicking on 'Proceed to Checkout' button to go to the checkout page"}
{"action": "terminate", "description": "Intent completed - successfully reached checkout page with highest rated meat substitute product within budget"}
\end{minted}

\end{document}